\documentclass{article}

\usepackage{microtype}
\usepackage{graphicx}
\usepackage{subfigure}
\usepackage{booktabs} 

\usepackage{hyperref}

\usepackage[accepted]{icml2023}

\usepackage{amsmath}
\usepackage{amssymb}
\usepackage{pifont}
\usepackage{mathtools}
\usepackage{amsthm}
\usepackage{wrapfig}

\usepackage[capitalize,noabbrev]{cleveref}

\usepackage{times}
\usepackage{latexsym}
\usepackage{amsmath}
\usepackage{fontawesome}
\usepackage[T1]{fontenc}

\usepackage{arydshln}

\usepackage[utf8]{inputenc}
 
\usepackage{microtype}
\usepackage{booktabs}
\usepackage{multicol}
\usepackage{multirow}
\usepackage{fancyhdr}
\usepackage{lastpage}
\usepackage{marvosym}
\usepackage{subfigure}
\usepackage{graphicx}
\usepackage{colortbl}
\usepackage{xcolor}
\usepackage{colortbl}
\usepackage{bm}

\newenvironment{itemize*}%
 {\leftmargini=10pt\begin{itemize}%
  \setlength{\itemsep}{0pt}%
  \setlength{\parskip}{0pt}%
  }%
 {\end{itemize}}
\newenvironment{enumerate*}%
 {\begin{enumerate}%
  \setlength{\itemsep}{0pt}%
  \setlength{\parskip}{0pt}}%
 {\end{enumerate}}

\definecolor{palepink}{rgb}{0.98, 0.85, 0.87}
\definecolor{a01}{rgb}{0.63,0.79,0.96}
\definecolor{b01}{rgb}{1,0.71,0.51}
\definecolor{c01}{rgb}{0.55,0.9,0.63}
\definecolor{d01}{rgb}{0.98,0.69,0.89}
\definecolor{d03}{rgb}{0.73,0.95,0.94}

\definecolor{a01_dm}{rgb}{0.19,0.54,0.91}
\definecolor{b01_dm}{rgb}{0.97,0.48,0.15}
\definecolor{c01_dm}{rgb}{0.12,0.93,0.3}
\definecolor{d01_dm}{rgb}{0.93,0.21,0.72}
\definecolor{d03_dm}{rgb}{0.15,0.61,0.61}

\definecolor{lightgreen}{rgb}
{0.95,0.97,0.93}
\newcolumntype{a}{>{\columncolor{lightgreen}}c}

\theoremstyle{plain}

\theoremstyle{definition}

\theoremstyle{remark}

\usepackage[textsize=tiny]{todonotes}

\icmltitlerunning{GPTScore: Evaluate as You Desire}

\begin{document}

\twocolumn[
\icmltitle{GPTScore: Evaluate as You Desire}




\begin{icmlauthorlist}
\icmlauthor{Jinlan Fu}{NUS}
\icmlauthor{See-Kiong Ng}{NUS}
\icmlauthor{Zhengbao Jiang}{CMU}
\icmlauthor{Pengfei Liu}{CMU}

\end{icmlauthorlist}

\icmlaffiliation{NUS}{National University of Singapore}
\icmlaffiliation{CMU}{Carnegie Mellon University}

\icmlcorrespondingauthor{Jinlan Fu}{jinlanjonna@gmail.com}
\icmlcorrespondingauthor{Pengfei Liu}{pliu3@cs.cmu.edu}

\icmlkeywords{Machine Learning, ICML}

\vskip 0.3in
]



\printAffiliationsAndNotice{}  


\begin{abstract}
Generative Artificial Intelligence (AI) has enabled the development of sophisticated models that are capable of producing high-caliber text, images, and other outputs through the utilization of large pre-trained models.
Nevertheless, assessing the quality of the generation is an even more arduous task than the generation itself, and this issue has not been given adequate consideration recently.
This paper proposes a novel evaluation framework, \textsc{GPTScore}, which utilizes the emergent abilities (e.g., zero-shot instruction) of \textbf{g}enerative \textbf{p}re-\textbf{t}rained models to \textbf{score} generated texts. 
There are 19 pre-trained models explored in this paper, ranging in size from 80M (e.g., FLAN-T5-small)  to 175B (e.g., GPT3).
Experimental results on four text generation tasks,  $22$ evaluation aspects, and corresponding $37$ datasets demonstrate that this approach can effectively allow us to achieve what one desires to evaluate for texts simply by natural language instructions.
This nature helps us overcome several long-standing challenges in text evaluation--how to achieve customized, multi-faceted evaluation without the need for annotated samples.
We make our code publicly available.~\footnote{\url{https://github.com/jinlanfu/GPTScore}}
\end{abstract}

\section{Introduction}

The advent of generative pre-trained models, such as GPT3 \cite{tom:gpt3}, has precipitated a shift from \textit{analytical} AI to \textit{generative} AI across multiple domains~\cite{sequoia2022}. Take text as an example: the use of a large pre-trained model with appropriate prompts \cite{liu2021pre} has achieved superior performance in tasks defined both in academia~\cite{sanh2021multitask} and scenarios from the real world~\cite{ouyang2022training}. While text generation technology is advancing rapidly, techniques for evaluating the quality of these texts lag far behind. This is especially evident in the following ways:

\begin{figure}[!th]
    \centering
    \includegraphics[width=0.97\linewidth]{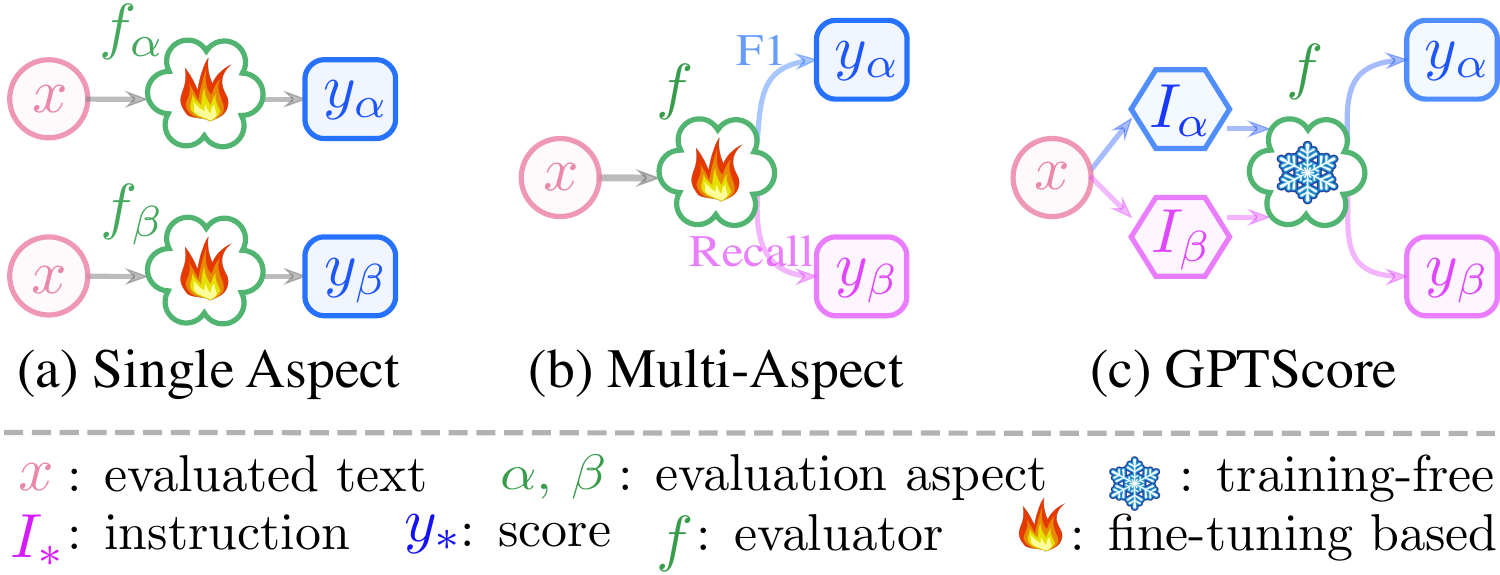}
    \vspace{-8pt}
    \caption{An overview of text evaluation approaches. 
}
	\label{fig:eval-overview}
\end{figure}

\noindent
(a) Existing studies evaluate text quality with limited aspects (e.g., semantic equivalence, fluency) (\autoref{fig:eval-overview}-(a)), which are usually customized prohibitively, making it harder for users to evaluate aspects \textit{as they need}~\cite{freitag:mqm}.
(b) A handful of studies have examined multi-aspect evaluation~\cite{bartscore2021yuan,QuestEval,zhong:unifiedEval} but have not given adequate attention to the definition of the evaluation aspect and the latent relationship among them. Instead, the evaluation of an aspect is either empirically bound with metric variants~\cite{bartscore2021yuan} or learned by supervised signals~\cite{zhong:unifiedEval}.
(c) Recently proposed evaluation methods~\cite{fed2020shikib,rei:comet,li:flowscore,zhong:unifiedEval} usually necessitate a complicated training procedure or costly manual annotation of samples  (\autoref{fig:eval-overview}-(a,b)), which makes it hard to use these methods in industrial settings due to the amount of time needed for annotation and training to accommodate a new evaluation demand from the user.

\begin{figure*}[ht]
    \centering
    \includegraphics[width=0.97\linewidth]{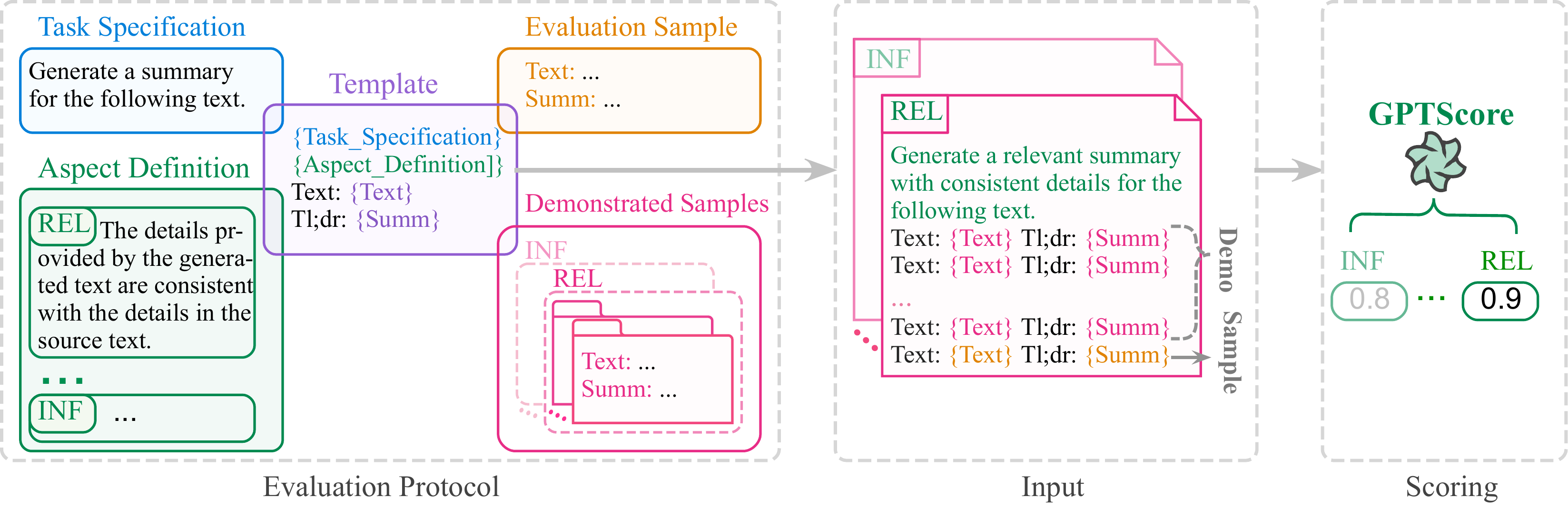}
    \vspace{-7pt}
    \caption{The framework of \textsc{GPTScore}. We include two evaluation aspects \textit{relevance (REL)} and \textit{informative (INF)} in this figure and use the evaluation of \textit{relevance (REL)} of the text summarization task to exemplify our framework.
    }
	\label{fig:framework}
\end{figure*}

In this paper, we demonstrated the talent of the super large pre-trained language model (e.g., GPT-3) in achieving multi-aspect, customized, and training-free evaluation  (\autoref{fig:eval-overview}-(c)). In essence, it skillfully uses the pre-trained model's zero-shot instruction~\cite{chung2022scaling}, and in-context learning~\cite{tom:gpt3,sewon:in-context} ability to deal with complex and ever-changing evaluation needs so as to solve multiple evaluation challenges that have been plagued for many years at the same time.

Specifically, given a text generated from a certain context, and desirable evaluation aspects (e.g., fluency), the high-level idea of the proposed framework is that the higher-quality text of a certain aspect will be more likely generated than unqualified ones based on the given context, where the ``likely'' can be measured by the conditional generation probability.
As illustrated in \autoref{fig:framework}, to capture users' true desires, an \textbf{\textit{evaluation protocol}} will be initially established based on (a) the \textit{task specification}, which typically
outlines how the text is generated (e.g., generate a response for a human based on the conversation.) (b) \textit{aspect definition} that documents the details of desirable evaluation aspects (e.g., the response should be intuitive to understand). Subsequently, each evaluation sample will be presented with the evaluated protocol with optionally moderate exemplar samples, which could facilitate the model's learning.
Lastly, a large \textbf{g}enerative \textbf{p}re-\textbf{t}rained model will be used to calculate how likely the text could be generated based on the above evaluation protocol, thus giving rise to our model's name: \textsc{GPTScore}.
Given the plethora of pre-trained models, we instantiate our framework with different backbones: GPT2~\cite{radford2019language}, OPT~\cite{zhang2022opt}, FLAN~\cite{chung2022scaling}, and GPT3 (instruction-based~\cite{ouyang2022training}) due to their superior capacity for \textit{zero-shot instruction} and their aptitude for \textit{in-context learning}.

Experimentally, we ran through almost all common natural language generation tasks in NLP, and the results showed the power of this new paradigm.
The main observations are listed as follows:
(1) Evaluating texts with generative pre-training models can be more reliable when instructed by the definition of \textit{task} and \textit{aspect}, providing a degree of flexibility to accommodate various evaluation criteria. Furthermore, incorporating exemplified samples with in-context learning will further enhance the process.
(2) Different evaluation aspects exhibit certain correlations. Combining definitions with other highly correlated aspects can improve evaluation performance.
(3) The performance of \texttt{GPT3-text-davinci-003}, which is tuned based on human feedback, is inferior to \texttt{GPT3-text-davinci-001} in the majority of the evaluation settings, necessitating deep explorations on the working mechanism of human feedback-based instruction learning (e.g., when it will fail).

\section{Preliminaries}

\subsection{Text Evaluation}
Text evaluation aims to assess the quality of hypothesis text $\bm{h}$ in terms of certain aspect $a$ (e.g., fluency), which is either measured manually with different protocols~\cite{nenkova-passonneau-2004-evaluating,realsumm2020manik,fabbri:summeval,liu2022revisiting} or quantified by diverse automated metrics~\cite{rouge2004lin,bleu2002kishore,moverscore2019wei,bertscore2020tianyi,bartscore2021yuan}.
\begin{align}
    y = f(\bm{h}, a, \mathcal{S})
\end{align}
where (1) $\bm{h}$ represents the text to be evaluated (hypothesis text, e.g., generated summary in text summarization task).
    (2) $a$ denotes the evaluation aspect (e.g., fluency). 
    (3) $\mathcal{S}$ is a collection of additional texts that are optionally used based on different scenarios. For example, it could be a source document or a reference summary in the text summarization task.
    (4) Function $f(\cdot)$ could be instantiated as a human evaluation process or automated evaluation metrics.

\subsection{Meta Evaluation}
Meta evaluation aims to evaluate the reliability of automated metrics by calculating how well automated scores ($y_{\text{auto}}$) correlate with human judgment ($y_{\text{human}}$) using correlation functions $g(y_{\text{auto}}, y_{\text{human}})$ such as spearman correlation.
In this work, we adopt two widely-used correlation measures:
(1) \textbf{Spearman} correlation ($\rho$)~\cite{zar2005spearman} measures the monotonic relationship between two variables based on their ranked values. 
(2) \textbf{Pearson} correlation ($r$)~\cite{mukaka2012guide} measures the linear relationship based on the raw data values of two variables.

\subsection{Evaluation Strategy}
Evaluation strategies define different aggregation methods when we calculate the correlation scores.
Specifically, suppose that for each source text $\bm{s}_{i}, i \in [1, 2, \cdots, n]$ (e.g., documents in text summarization task or dialogue histories for dialogue generation task), there are $J$ system outputs $\bm{h}_{i,j}$, where $j \in [1, 2, \cdots, J]$. 
$f_{\text{auto}}$ is an automatic scoring function (e.g., ROUGE~\cite{rouge2004lin}), and $f_{\text{human}}$ is the gold human scoring function.
For a given evaluation aspect $a$, the meta-evaluation metric $F$ can be formulated as follows.

\paragraph{Sample-level} defines that a correlation value is calculated for each sample separately based on outputs of multiple systems, then averaged across all samples.
\begin{equation*}
\begin{split}
    F^{\text{sample}}_{f_{\text{auto}}, f_{\text{human}}} = \frac{1}{n} \sum\limits_{i=1}^{n} 
    \Big(g\big(
    \left[f_{\text{auto}}(\bm{h}_{i,1}), \cdots, f_{\text{auto}}(\bm{h}_{i,J}) \right], \\
    \left[f_{\text{human}}(\bm{h}_{i,1}), \cdots, f_{\text{human}}(\bm{h}_{i,J}) \right]
    \big) \Big),
\end{split}
\end{equation*}
where $g$ can be instantiated as Spearman or Pearson correlation.

\paragraph{Dataset-level} indicates that the correlation value is calculated on system outputs of all $n$ samples.
\begin{equation*}
\begin{split}
    F^{\text{data}}_{f_{\text{auto}}, f_{\text{human}}} = g\Big(
    \left[f_{\text{auto}}(\bm{h}_{1,1}), \cdots, f_{\text{auto}}(\bm{h}_{n, J})\right], \\
    \left[f_{\text{human}}(\bm{h}_{1,1}), \cdots, f_{\text{human}}(\bm{h}_{n, J})\right]
    \Big)
\end{split}
\end{equation*}
In this work, we select the evaluation strategy for a specific task based on previous works~\cite{bartscore2021yuan,Zhang:fineD}. 
We use the sample-level evaluation strategy for text summarization, data-to-text, and machine translation tasks.
For the dialogue response generation task, the dataset-level evaluation strategy is utilized.

\section{\textsc{GPTScore}}

\subsection{Generative Pre-trained Language Models}
Existing pre-trained language models could be classified into the following three categories:
(a) encoder-only models (e.g., BERT~\cite{devlin:bert}, RoBerta~\cite{liu:roberta}) that encode inputs with bidirectional attention; 
(b) encoder-decoder models (e.g., BART~\cite{bart:mike}, T5~\cite{colin:t5}) that encode inputs with bidirectional attention and generate outputs autoregressively; 
(c) decoder-only models (e.g., GPT2~\cite{radford2019language}, GPT3~\cite{tom:gpt3}, PaLM~\cite{chowdhery:palm}) that generate the entire text sequence autoregressively, where pre-trained models with decoding abilities (b, c) have caught much attention since they show impressive performance on zero-shot instruction and in-context learning.
Specifically, given a prompt text $\bm{x} = \{x_1, x_2, \cdots, x_n\}$, a generative pre-training language model can generate a textual continuation $\bm{y} = \{y_1, y_2, \cdots, y_m\}$ with the following generation probability:
\begin{equation*}
    p(\bm{y}|\bm{x}, \theta) = \prod_{t=1}^{m}p(y_t | \bm{y}_{<t}, \bm{x}, \theta)
\end{equation*}
\textbf{Emergent Ability}
Recent works progressively reveal a variety  of emergent abilities of generative pre-trained language models with appropriate tuning or prompting methods, such as in-context learning~\cite{sewon:in-context}, chain-of-thought reasoning~\cite{jason2022chain}, and zero-shot instruction~\cite{ouyang2022training}. One core commonality of these abilities is to allow for handling customized requirements with a few or even zero annotated examples.
It's the appearance of these abilities that allows us to re-invent a new way for text evaluation--evaluating from the textual description, which can achieve customizable, multi-faceted, and train-free evaluation.

\subsection{Generative Pretraining Score (GPTScore)}
The core idea of \textsc{GPTScore} is that a generative pre-training model will assign a higher probability of high-quality generated text following a given instruction and context.
In our method, the instruction is composed of the task description $d$ and the aspect definition $a$. Specifically, suppose that the text to be evaluated is $\bm{h} = \{h_1, h_2, \cdots, h_m\}$, the context information is $\mathcal{S}$ (e.g., source text or reference text), then \textsc{GPTScore} is defined as the following conditional probability:
\begin{equation*}
    \mathrm{GPTScore}(\bm{h}|d, a, \mathcal{S}) = \sum_{t=1}^{m} w_t \log p(h_t |\bm{h}_{<t}, T(d, a, \mathcal{S}), \theta),
\end{equation*}
where $w_t$ is the weight of the token at position $t$. In our work, we treat each token equally. $T(\cdot)$ is a prompt template that defines the evaluation protocol, which is usually task-dependent and specified manually through prompt engineering.

\begin{table*}[htb]
  \centering  \scriptsize  
   \renewcommand\tabcolsep{1.7pt}
    \begin{tabular}{llm{10cm}}
    \toprule
    \textbf{Aspect} & \textbf{Task} & \textbf{Definition} \\
    \midrule
    Semantic Coverage (COV) & Summ  & How many semantic content units from the reference text are covered by the generated text? \\
    Factuality (FAC) & Summ  & Does the generated text preserve the factual statements of the source text? \\
    Consistency (CON) & Summ, Diag & Is the generated text consistent in the information it provides? \\
    Informativeness (INF) & Summ, D2T, Diag & How well does the generated text capture the key ideas of its source text? \\
    Coherence (COH) & Summ, Diag & How much does the generated text make sense? \\
    Relevance (REL) & Diag, Summ, D2T & How well is the generated text relevant to its source text? \\
    Fluency (FLU) & Diag, Summ, D2T, MT & Is the generated text well-written and grammatical? \\
    Accuracy (ACC) & MT    & Are there inaccuracies, missing, or unfactual content in the generated text? \\
    Multidimensional  & \multirow{2}[0]{*}{MT} & \multicolumn{1}{l}{\multirow{2}[0]{*}{How is the overall quality of the generated text?}} \\
    Quality Metrics (MQM) &       &  \\
    Interest (INT) & Diag  & Is the generated text interesting? \\
    Engagement (ENG) & Diag  & Is the generated text engaging? \\
    Specific (SPE) & Diag  & Is the generated text generic or specific to the source text? \\
    Correctness (COR) & Diag  & Is the generated text correct or was there a misunderstanding of the source text? \\
    Semantically & \multirow{2}[0]{*}{Diag}  & \multicolumn{1}{l}{\multirow{2}[0]{*}{Is the generated text semantically appropriate?}} \\
    appropriate (SEM)  &        &    \\
    Understandability (UND) & Diag  & Is the generated text understandable? \\
    Error Recovery (ERR) & Diag  & Is the system able to recover from errors that it makes? \\
    Diversity (DIV) & Diag  & Is there diversity in the system responses? \\
    Depth (DEP) & Diag  & Does the system discuss topics in depth? \\
    Likeability (LIK) & Diag  & Does the system display a likeable personality? \\
    Flexibility (FLE) & Diag  & Is the system flexible and adaptable to the user and their interests? \\
    Inquisitiveness (INQ) & Diag  & Is the system inquisitive throughout the conversation? \\
    \bottomrule
    \end{tabular}%
        \vspace{-6pt}
      \caption{
      The definition of aspects evaluated in this work.
      \textit{Semantic App.} denotes \textit{semantically appropriate} aspect. \textit{Diag}, \textit{Summ}, \textit{D2T}, and \textit{MT} denote the \textit{dialogue response generation}, \textit{text summarization}, \textit{data to text} and \textit{machine translation}, respectively.
      }  
  \label{tab:asp-define}%
\end{table*}%

\textbf{Few-shot with Demonstration} \quad
The generative pre-trained language model can better perform tasks when prefixed with a few annotated samples (i.e., demonstrations). Our proposed framework is flexible in supporting this by extending the prompt template $T$ with demonstrations. 

\textbf{Choice of Prompt Template} \quad
Prompt templates define how task description, aspect definition, and context are organized. Minging desirable prompts itself is a non-trivial task and there are extensive research works there~\cite{liu2021pre,fu2022polyglot}. In this work, for the GPT3-based model, we opt for prompts that are officially provided by OpenAI.\footnote{\url{https://beta.openai.com/examples}} For instruction-based pre-trained models, we use prompts from NaturalInstruction~\cite{wang2022super} since it's the main training source for those instruction-based pre-train models.
Taking the evaluation of the fluency of the text summarization task as an example, based on the prompt provided by OpenAI,\footnote{\url{https://beta.openai.com/examples/default-tldr-summary}} the task prompt is  ``\{Text\} Tl;dr \{Summary\}'', the definition of fluency is ``Is the generated text well-written and grammatical?'' (in \autoref{tab:asp-define}), and then the final prompt template is ``\texttt{Generate a fluent and grammatical summary for the following text: \{Text\} Tl;dr \{Summary\}}'', where demonstrations could be introduced by repeating instantiating ``\texttt{\{Text\} Tl;dr \{Summary\}}''
In \autoref{sec:prompt}, we list the prompts for various aspects of all tasks studied in this work and leave a more comprehensive exploration on prompt engineering as a future work.

\textbf{Selection of Scoring Dimension} \quad
\textsc{GPTScore} exhibits different variants in terms of diverse choices of texts being calculated. 
For example, given a generated hypothesis, we can calculate \textsc{GPTScore} either based on the source text (i.e., \textit{src->hypo}, $p(\text{hypo}|\text{src})$) or based on the gold reference (i.e., \textit{ref->hypo}, $p(\text{hypo}|\text{ref})$).
In this paper, the criteria for choosing \textsc{GPTScore} variants are mainly designed to align the protocol of human judgments~\cite{liu2022revisiting} that are used to evaluate the reliability of automated metrics.
We will detail this based on different human judgment datasets in the experiment section.

\section{Experimental Settings}

\subsection{Tasks, Datasets, and Aspects}
\label{sec:asp}
To achieve a comprehensive evaluation, in this paper, we cover a broad range of natural language generation tasks:
\textit{Dialogue Response Generation}, \textit{Text Summarization}, \textit{Data-to-Text}, and \textit{Machine Translation}, which involves $37$ datasets and $22$ evaluation aspects in total.
\autoref{tab:task-data} summarizes the tasks, datasets, and evaluation aspects considered by each dataset. 
The definition of different aspects can be found in \autoref{tab:asp-define}. More detailed illustrations about the datasets can be found in \autoref{sec:app-task-asp}.

(1) \textbf{Dialogue Response Generation} aims to automatically generate an engaging and informative response based on the dialogue history. Here, we choose to use the \texttt{FED}~\cite{fed2020shikib} datasets and consider both turn-level and dialogue-level evaluations.
(2) \textbf{Text Summarization} is a task of automatically generating informative and fluent summary for a given long text. Here, we consider the following four datasets, \texttt{SummEval}~\cite{realsumm2020manik}, 
\texttt{REALSumm}~\cite{realsumm2020manik}, \texttt{NEWSROOM}~\cite{newsroom2018max}, and \texttt{QAGS\_XSUM}~\cite{qags20alex}, covering 10 aspects.
(3) \textbf{Data-to-Text} aims to automatically generate a fluent and factual description for a given table.
Our work considered \texttt{BAGEL}~\cite{bagel2010francois} and \texttt{SFRES}~\cite{sfres2015tsung} datasets.
(4) \textbf{Machine Translation} aims to translate a sentence from one language to another. We consider a subdatasets of Multidimensional Quality Metrics (MQM)~\cite{freitag:mqm}, namely, \texttt{MQM-2020} (Chinese->English).

\subsection{Scoring Models}
     \textbf{ROUGE}~\cite{rouge2004lin} is a popular automatic generation evaluation metric. We consider three variants ROUGE-1, ROUGE-2, and ROUGE-L.
     \textbf{PRISM}~\cite{prism:brian} is a reference-based evaluation method designed for machine translation with pre-trained paraphrase systems.
     \textbf{BERTScore}~\cite{bertscore2020tianyi} uses contextual representation from BERT to calculate the similarity between the generated text and the reference text. 
     \textbf{MoverScore}~\cite{moverscore2019wei} considers both contextual representation and Word Mover's Distance (WMD, \cite{wmd:kusner})
     \textbf{DynaEval}~\cite{zhang:dynaeval} is a unified automatic evaluation framework for dialogue response generation tasks on the turn level and dialogue level.
    \textbf{BARTScore}~\cite{bartscore2021yuan} is a text-scoring model based on BART~\cite{bart:mike} without fine-tuning.    
     \textbf{BARTScore+CNN}~\cite{bartscore2021yuan} is based on BART fine-tuned on the CNNDM dataset~\cite{cnndm:karl}. 
    \textbf{BARTScore+CNN+Para}~\cite{bartscore2021yuan} is based on BART fine-tuned on CNNDM and Paraphrase2.0~\cite{paraphrase:edward}. 
     \textbf{\textsc{GPTScore}} is our evaluation method, which is designed based on different pre-trained language models. Specifically, we considered GPT3, OPT, FLAN-T5, and GPT2 in this work. Five variants are explored for each framework. For a fair comparison with the decoder-only model, such as GPT3 and OPT, only four variant models of GPT2 with a parameter size of at least 350M are considered. \autoref{tab:model-params} shows all model variants we used in this paper and their number of parameters.

\begin{table}[htb]
  \centering \footnotesize
    \begin{tabular}{lala}
    \toprule
    \textbf{GPT3} & \textbf{Param.} & \textbf{OPT} & \textbf{Param.} \\
    \midrule
    text-ada-001 & 350M  & OPT350M & 350M \\
    text-babbage-001 & 1.3B  & OPT-1.3B & 1.3B \\
    text-curie-001 & 6.7B  & OPT-6.7B & 6.7B \\
    text-davinci-001 & 175B  & OPT-13B & 13B \\
    text-davinci-003 & 175B  & OPT-66B & 66B \\
    \midrule
    \textbf{FLAN-T5} & \textbf{Param.} & \textbf{GPT2} & \textbf{Param.} \\
   \midrule
    FT5-small & 80M   & GPT2-M & 355M \\
    FT5-base & 250M  & GPT2-L & 774M \\
    FT5-L & 770M  & GPT2-XL & 1.5B \\
    FT5-XL & 3B    & GPT-J-6B & 6B \\
    FT5-XXL & 11B   &       &  \\
    \bottomrule
    \end{tabular}%
    \vspace{-6pt}
    \caption{Pre-trained backbones used in this work.}
  \label{tab:model-params}%
\end{table}%

\subsection{Scoring Dimension}
Specifically, 
(1) For aspects \texttt{INT}, \texttt{ENG}, \texttt{SPC}, \texttt{REL}, \texttt{COR}, \texttt{SEM}, \texttt{UND}, and \texttt{FLU} of FED-Turn datasets from the open domain dialogue generation task, we choose the \textit{src->hypo} variant since the human judgments of the evaluated dataset (i.e., FED-Turn) are also created based on the source.
(2)
For aspects \texttt{COH}, \texttt{CON}, and \texttt{INF} from SummEval and Newsroom, since data annotators labeled the data based on source and hypothesis texts, we chose \textit{src->hypo} for these aspects.
(3) 
For aspects \texttt{INF}, \texttt{NAT}, and \texttt{QUA} from the data-to-text task, we choose \textit{src->hypo}.  Because the source text of the data-to-text task is not in the standard text format, which will be hard to handle by the scoring function.
(4) 
For aspects \texttt{ACC}, \texttt{FLU}, and \texttt{MQM} from the machine translation task, we also choose \textit{src->hypo}. Because the source text of the machine translation is a different language from the translated text (hypo). In this work, we mainly consider the evaluation of the English text. In the future, we can consider designing a scoring function based on BLOOM~\cite{google:bloom} that can evaluate texts in a cross-lingual setting.

\subsection{Evaluation Dataset Construction}
Unlike previous works~\cite{carp:2021,SEScore2:2022,wenda:2022,robust:2022} that only consider the overall text quality, we focus on evaluating multi-dimensional text quality.
In this work, we studied 37 datasets according to 22 evaluation aspects. Due to the expensive API cost of GPT3, we randomly extract and construct sub-datasets for meta-evaluation.
For the MQM dataset, since many aspects of samples lack human scores, we extract samples with human scores in \texttt{ACC}, \texttt{MQM}, and \texttt{FLU} as much as possible.

\section{Experiment Results}

In this work, we focus on exploring whether language models with different structures and sizes can work in the following three scenarios. (a) \textbf{vanilla (VAL)}: with non-instruction and non-demonstration; (b) \textbf{instruction (IST)}: with instruction and non-demonstration; (c) \textbf{instruction+demonstration (IDM)}: with instruction and demonstration.

\textbf{Significance Tests}
To examine the reliability and validity of the experiment results, we conducted the significance test based on bootstrapping.\footnote{\url{https://en.wikipedia.org/wiki/Bootstrapping_(statistics)}}
Our significance test is to check
(1) whether the performance of IST (IDM) is significantly better than VAL, and values achieved with the IST (IDM) settings will be marked $\dag$ if it passes the significant test (p-value <0.05).
(2) whether the performance of IDM is significantly better than IST, if yes, mark the value with IDM setting with $\ddag$.

\textbf{Average Performance}
Due to space limitations, we keep the average performance of GPT3-based, GPT2-based, OPT-based, and FT5-based models. The full results of various variants can be found in \autoref{sec:exp-res-full}.

\begin{table}[htb]
  \centering \footnotesize
   \renewcommand\tabcolsep{0.5pt}
   \renewcommand\arraystretch{0.93}  
    \begin{tabular}{lcccccccccc}
    \toprule
    \multirow{3}[4]{*}{Model} & \multicolumn{8}{c}{SummEval}                                  & \multicolumn{2}{c}{RSumm} \\
\cmidrule{2-11}          & \multicolumn{2}{c}{COH} & \multicolumn{2}{c}{CON} & \multicolumn{2}{c}{FLU} & \multicolumn{2}{c}{REL} & \multicolumn{2}{c}{COV} \\
\cmidrule(lr){2-3}\cmidrule(lr){4-5}\cmidrule(lr){6-7}\cmidrule(lr){8-9}\cmidrule(lr){10-11}
          & VAL   & IST   & VAL   & IST   & VAL   & IST   & VAL   & IST   & VAL   & IST \\
    \midrule
    ROUGE-1 & 14.1  & -     & 20.8  & -     & 14.8  & -     & 26.2  & -     & \textbf{46.4} & - \\
    ROUGE-2 & 9.1   & -     & 17.2  & -     & 12.0  & -     & 17.4  & -     & 37.3  & - \\
    ROUGE-L & 12.9  & -     & 19.8  & -     & 17.6  & -     & 24.7  & -     & 45.1  & - \\
    BERTSc & 25.9  & -     & 19.7  & -     & 23.7  & -     & 34.7  & -     & 38.4  & - \\
    MoverSc & 11.5  & -     & 18.0  & -     & 15.7  & -     & 24.8  & -     & 34.4  & - \\
    PRISM & 26.5  & -     & 29.9  & -     & 26.1  & -     & 25.2  & -     & 32.3  & - \\
    BARTSc & 29.7  & -     & 30.8  & -     & 24.6  & -     & 28.9  & -     & 43.1  & - \\
    +CNN  & 42.5  & -     & 35.8  & -     & 38.1  & -     & \textbf{35.9} & -     & 42.9  & - \\
    +CNN+Pa & \textbf{42.5} & -     & \textbf{37.0} & -     & \textbf{40.5} & -     & 33.9  & -     & 40.9  & - \\
    \midrule
    GPT3-a01 & 39.3  & $\text{39.8}^{\dag}$ & 39.7  & $\text{40.5}^{\dag}$ & 36.1  & 35.9  & 28.2  & 27.6  & 29.5  & $\text{29.8}^{\dag}$ \\
    GPT3-b01 & 42.7  & $\text{\textbf{45.2}}^{\dag}$ & 41.0  & $\text{41.4}^{\dag}$ & 37.1  & $\text{39.1}^{\dag}$ & 32.0  & $\text{33.4}^{\dag}$ & 35.0  & $\text{35.2}^{\dag}$ \\
    GPT3-c01 & 41.3  & 40.8  & 44.6  & $\text{45.1}^{\dag}$ & 38.9  & $\text{39.5}^{\dag}$ & 31.6  & $\text{33.2}^{\dag}$ & \textbf{36.1} & $\text{\textbf{45.1}}^{\dag}$ \\
    GPT3-d01 & 40.0  & 40.1  & \textbf{46.6} & $\text{\textbf{47.5}}^{\dag}$ & 40.5  & $\text{\textbf{41.0}}^{\dag}$ & 32.4  & $\text{34.3}^{\dag}$ & 36.0  & 33.9 \\
    GPT3-d03 & \textbf{43.7} & 43.4  & 45.2  & 44.9  & \textbf{41.1} & 40.3  & \textbf{36.3} & $\text{\textbf{38.1}}^{\dag}$ & 35.2  & $\text{38.0}^{\dag}$ \\
    \midrule
    GPT2-M & 36.0  & $\text{39.2}^{\dag}$ & 34.6  & $\text{35.3}^{\dag}$ & 28.1  & $\text{30.7}^{\dag}$ & 28.3  & 28.3  & 41.8  & $\text{43.3}^{\dag}$ \\
    GPT2-L & \textbf{36.4} & $\text{39.8}^{\dag}$ & 33.7  & $\text{34.4}^{\dag}$ & 29.4  & $\text{31.5}^{\dag}$ & 27.8  & $\text{28.1}^{\dag}$ & 39.6  & $\text{41.3}^{\dag}$\\
    GPT2-XL & 35.3  & $\text{\textbf{39.9}}^{\dag}$ & 35.9  & $\text{36.1}^{\dag}$ & 31.2  & $\text{33.1}^{\dag}$ & 28.1  & 28.0  & 40.4  & $\text{41.0}^{\dag}$ \\
    GPT-J-6B & 35.5  & $\text{39.5}^{\dag}$ & \textbf{42.7} & $\text{\textbf{42.8}}^{\dag}$ & \textbf{35.5} & $\text{\textbf{37.4}}^{\dag}$ & \textbf{31.5} & $\text{\textbf{31.9}}^{\dag}$ & \textbf{42.8} & $\text{\textbf{43.7}}^{\dag}$ \\
    \midrule
    OPT350m & 33.4  & $\text{37.6}^{\dag}$ & 34.9  & $\text{35.5}^{\dag}$ & 29.6  & $\text{31.4}^{\dag}$ & 29.5  & 28.6  & 40.2  & $\text{42.3}^{\dag}$ \\
    OPT-1.3B & 35.0  & $\text{\textbf{37.8}}^{\dag}$ & 40.0  & $\text{42.0}^{\dag}$ & 33.6  & $\text{35.9}^{\dag}$ & 33.5  & $\text{34.2}^{\dag}$ & 42.0  & 39.7 \\
    OPT-6.7B & \textbf{35.7} & $\text{36.8}^{\dag}$ & 42.1  & $\text{\textbf{45.7}}^{\dag}$ & 35.5  & $\text{37.6}^{\dag}$ & \textbf{35.4} & \textbf{35.4} & 38.0  & $\text{41.9}^{\dag}$ \\
    OPT-13B & 33.5  & $\text{34.7}^{\dag}$ & 42.5  & $\text{45.2}^{\dag}$ & 35.6  & $\text{37.3}^{\dag}$ & 33.6  & 33.9  & 37.6  & $\text{41.0}^{\dag}$ \\
    OPT-66B & 32.0  & $\text{35.9}^{\dag}$ & \textbf{44.0} & $\text{45.3}^{\dag}$ & \textbf{36.3} & $\text{\textbf{38.0}}^{\dag}$ & 33.4  & $\text{33.7}^{\dag}$& \textbf{40.3} & $\text{41.3}^{\dag}$ \\
    \midrule
    FT5-small & 35.0  & $\text{35.4}^{\dag}$ & 37.0  & $\text{38.0}^{\dag}$ & 35.6  & 34.7  & 27.3  & $\text{28.0}^{\dag}$ & 33.6  & $\text{35.7}^{\dag}$ \\
    FT5-base & 39.2  & $\text{39.9}^{\dag}$ & 36.7  & $\text{37.2}^{\dag}$ & 37.3  & 36.5  & 29.5  & $\text{31.2}^{\dag}$ & 36.7  & $\text{38.6}^{\dag}$ \\
    FT5-L & 42.3  & $\text{45.1}^{\dag}$ & 41.0  & $\text{42.5}^{\dag}$ & 39.3  & $\text{41.6}^{\dag}$ & 31.2  & $\text{\textbf{35.3}}^{\dag}$ & 31.4  & $\text{39.3}^{\dag}$ \\
    FT5-XL & \textbf{42.8} & $\text{\textbf{47.0}}^{\dag}$ & 41.0  & $\text{43.6}^{\dag}$ & 39.7  & $\text{42.1}^{\dag}$ & 31.4  & $\text{34.4}^{\dag}$ & 34.8  & $\text{\textbf{43.8}}^{\dag}$ \\
    FT5-XXL & 42.1  & $\text{45.6}^{\dag}$ & \textbf{43.7} & \textbf{43.8} & \textbf{39.8} & $\text{\textbf{42.4}}^{\dag}$ & \textbf{32.8} & $\text{34.3}^{\dag}$ & \textbf{40.2} & $\text{41.1}^{\dag}$ \\
    \midrule
   \rowcolor{palepink}  Avg. & 38.0  & 40.2  & 40.4  & 41.4  & 35.8  & 37.2  & 31.3  & 32.2  & 37.4  & 39.8 \\
    \bottomrule
    \end{tabular}
    \vspace{-8pt}
  \caption{Spearman correlation  of different aspects on text summarization datasets. 
    {VAL} and IST is the abbreviation of vanilla and instruction, respectively.
      Values with $\dag$ denote the evaluator with instruction significantly outperforms with vanilla. Values in bold are the best performance in a set of variants (e.g., GPT3 family). 
      }
  \label{tab:res-summ}
\end{table}

\subsection{Text Summarization}
\begin{figure}[htb]
  \centering \footnotesize
  \renewcommand\tabcolsep{0.5pt}
  \renewcommand\arraystretch{0.93}  
    \begin{tabular}{ccccccccccccccc}
    \toprule
    \multicolumn{12}{c}{SummEval}                                             & \multicolumn{3}{c}{RSumm} \\
    \cmidrule(lr){1-12}\cmidrule(lr){13-15}
    \multicolumn{3}{c}{COH} & \multicolumn{3}{c}{CON} & \multicolumn{3}{c}{FLU} & \multicolumn{3}{c}{REl} & \multicolumn{3}{c}{COV} \\
    \midrule
    \multicolumn{3}{c}{\multirow{5}[2]{*}{\includegraphics[scale=0.3]{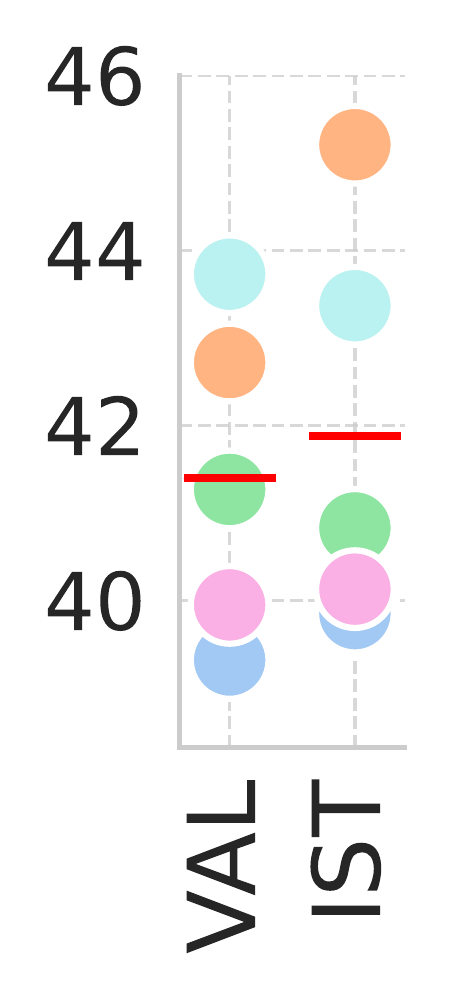}}} 
    & \multicolumn{3}{c}{\multirow{5}[2]{*}{\includegraphics[scale=0.3]{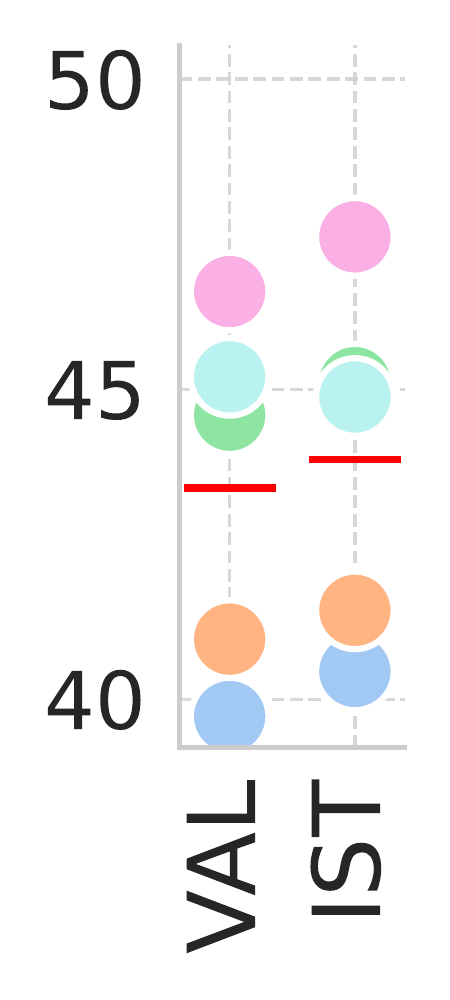}}} 
    & \multicolumn{3}{c}{\multirow{5}[2]{*}{\includegraphics[scale=0.3]{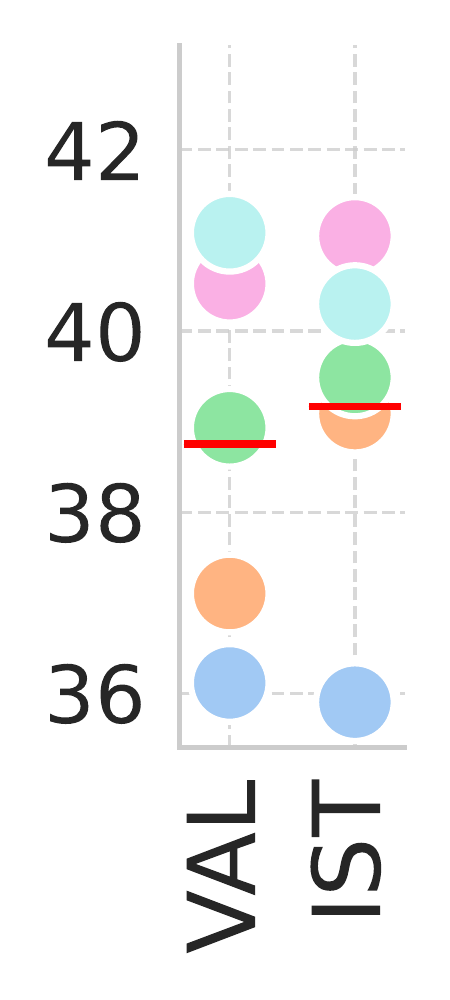}}} 
    & \multicolumn{3}{c}{\multirow{5}[2]{*}{\includegraphics[scale=0.3]{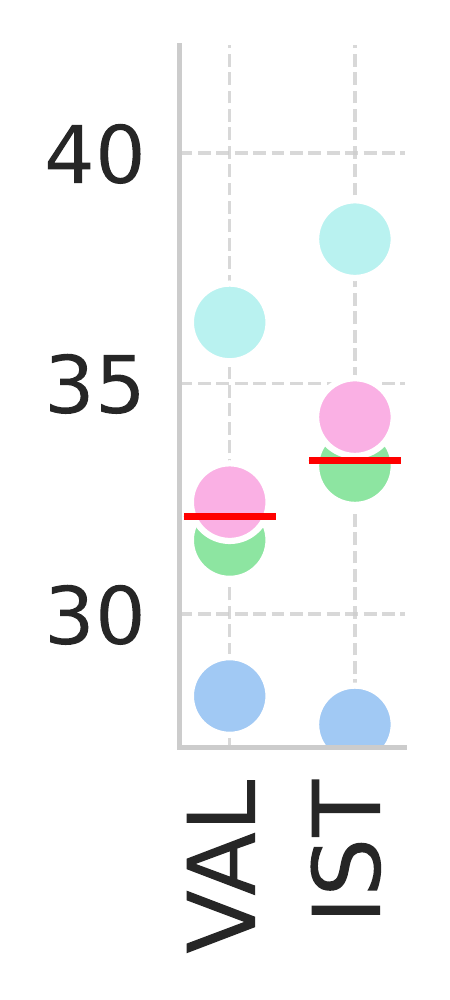}}} 
    & \multicolumn{3}{c}{\multirow{5}[2]{*}{\includegraphics[scale=0.3]{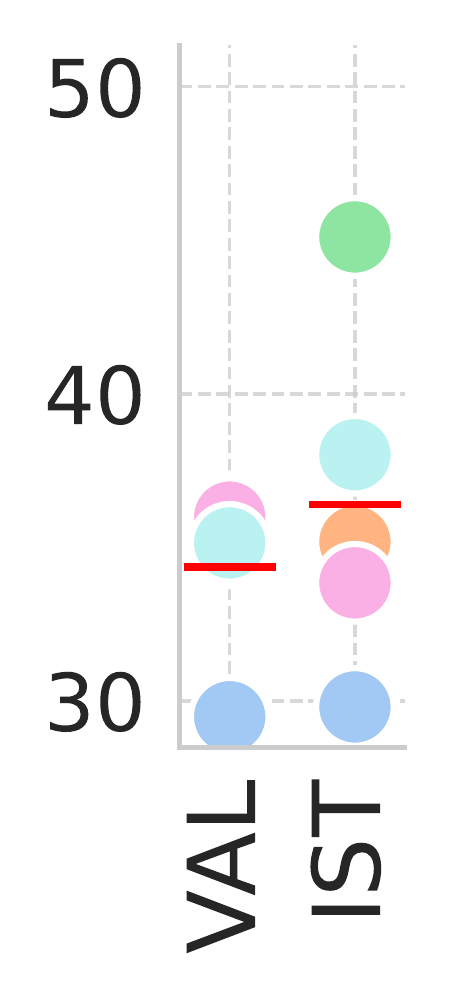}}}  \\ \\ \\ \\ \\ \\ \\  \\ \\ 
    \midrule
    \multicolumn{12}{c}{NEWSROOM}                                             & \multicolumn{3}{c}{QXSUM} \\
    \cmidrule(lr){1-12}\cmidrule(lr){13-15}
    \multicolumn{3}{c}{COH} & \multicolumn{3}{c}{FLU} & \multicolumn{3}{c}{REL} & \multicolumn{3}{c}{INF} & \multicolumn{3}{c}{FAC} \\
    \midrule
    \multicolumn{3}{c}{\multirow{5}[2]{*}{\includegraphics[scale=0.3]{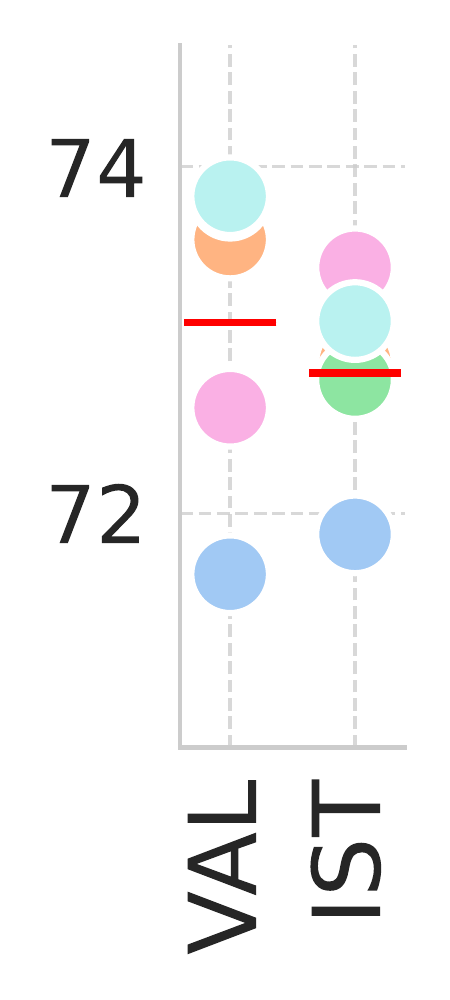}}} 
    & \multicolumn{3}{c}{\multirow{5}[2]{*}{\includegraphics[scale=0.3]{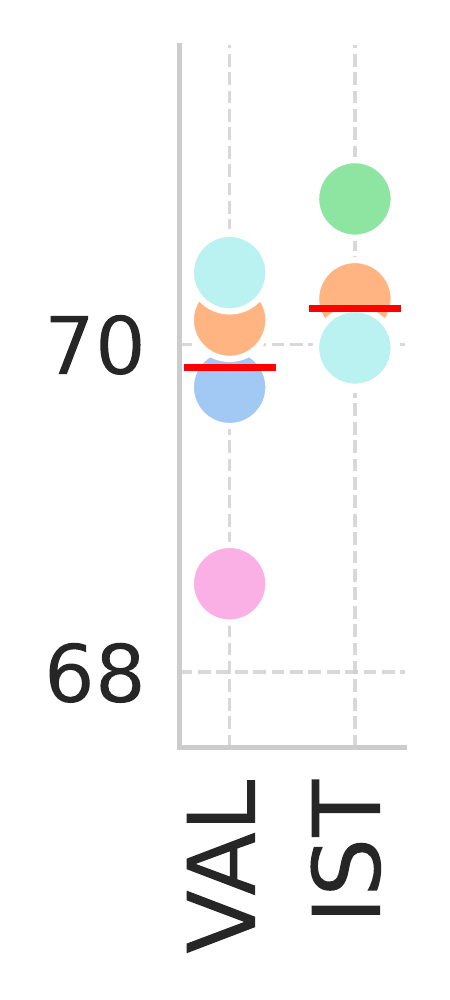}}} 
    & \multicolumn{3}{c}{\multirow{5}[2]{*}{\includegraphics[scale=0.3]{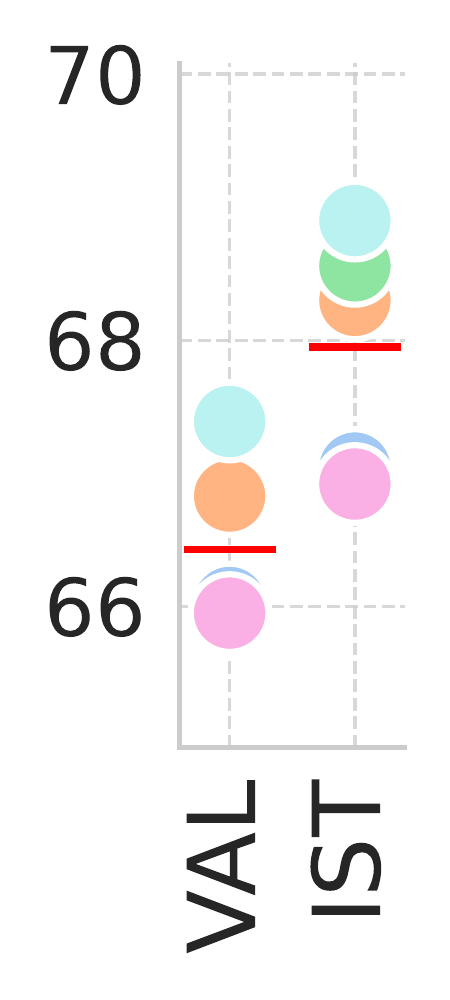}}} 
    & \multicolumn{3}{c}{\multirow{5}[2]{*}{\includegraphics[scale=0.3]{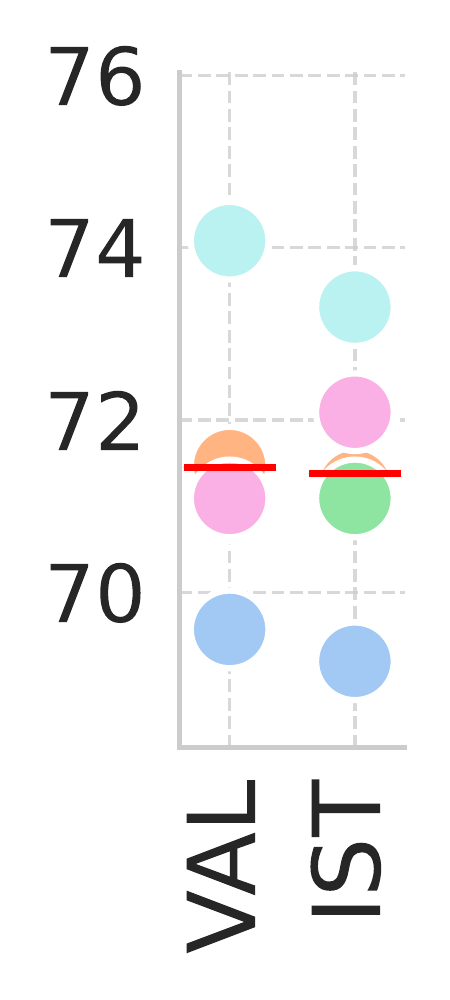}}} 
    & \multicolumn{3}{c}{\multirow{5}[2]{*}{\includegraphics[scale=0.3]{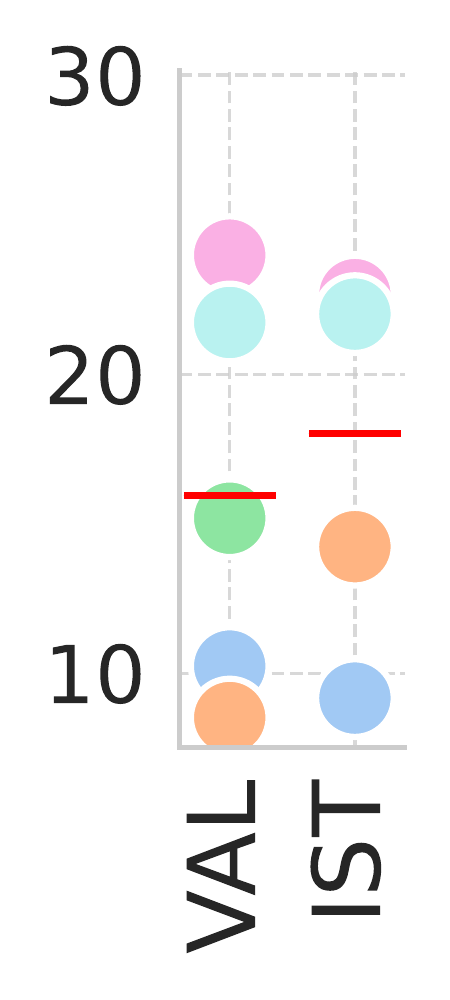}}}  \\ \\ \\ \\ \\ \\ \\  \\ \\ 
    \bottomrule
    \end{tabular}
  \vspace{-6pt}
  \caption{Experimental results for GPT3-based variants in text summarization task. Here, 
 blue,
 orange,
 green, 
 pink, 
 and 
 cyan dot denote that \textsc{GPTScore} is built based on 
 a01 (\textcolor{a01}{\faCircle}), 
 b01 (\textcolor{b01}{\faCircle}), 
 c01 (\textcolor{c01}{\faCircle}), 
 d01 (\textcolor{d01}{\faCircle}), and 
 d03 (\textcolor{d03}{\faCircle}), respectively. The
 red lines (\textcolor{red}{\textbf{---}}) denote the average performance of GPT3-based variants. 
 }
   \label{fig:res-summ-gpt3}%
\end{figure}

The evaluation results of $28$ (9 baseline models (e.g., ROUGE-1) and 19 variants of GPTScore (e.g., GPT3-d01)) scoring functions for the text summarization task on SummEval and RealSumm datasets are shown in \autoref{tab:res-summ}. Due to the space limitation, we move the performance of the NEWSROOM and QXSUM datasets to the \autoref{sec:exp-res-full}.    
\autoref{fig:res-summ-gpt3} shows the evaluation results of five GPT3 variant models on four text summarization datasets, where QXSUM uses the Pearson correlation and other datasets use the Spearman correlation metric.
The main observations are summarized as follows:

    (1) \textbf{Evaluator with instruction significantly improves the performance} (values with $\dag$ in \autoref{tab:res-summ}). What's more, small models with instruction demonstrate comparable performance 
    to supervised learning models. For example, OPT350m, FT5-small, and FT5-base outperform BARTScore+CNN on the \texttt{CON} aspect when using the instructions.
    (2) \textbf{The benefit from instruction is more stable for the decoder-only models.} In \autoref{tab:res-summ}, the average Spearman score of both the GPT2 and OPT models, 9 out of 10 aspects are better than the vanilla setting (VAL)  by using instruction (IST), while the equipment of instruction (IST) to the encoder-decoder model of FT5 on the NEWSROOM dataset fails to achieve gains.
    (3) As for the GPT3-based models, \textbf{(a) the performance of GPT3-d01 is barely significantly better than GPT3-c01}, which tries to balance power and speed. (b) \textbf{GPT3-d03 performs better than GPT3-d01 significantly.} We can observe these conclusions from \autoref{fig:res-summ-gpt3}, and both conclusions have passed the significance test at $p<0.05$.

\subsection{Machine Translation}

The average sample-level Spearman ($\rho$) scores of GPT3-based, GPT2-based, OPT-based, and FT5-based models on the MQM-2020 machine translation dataset are shown in \autoref{tab:res-mqm}, where values with $\dag$ denote that the evaluator equipped with IST (or IDM) significantly outperforms the VAL setting, and 
$\ddag$ indicate that the evaluator equipped with IDM (the combination of IST and DM) significantly outperforms the IST setting. 
The Spearman correlations for the GPT3-based variants are shown in \autoref{fig:res-mqm-gpt3}. 
For the full evaluation results of $28$ models (including $9$ baseline scoring models, such as ROUGE-1) can be found in \autoref{tab:mqm2020-full}.
Following~\citet{prism:brian} and ~\citet{bartscore2021yuan}, we treat the evaluation of machine translation as the paraphrasing task. The main observations are listed as follows:

    (1) \textbf{The introduction of instruction (IST) significantly improve the performance in three different aspects of \texttt{ACC}, \texttt{FLU}, and \texttt{MQM}.} In \autoref{tab:res-mqm}, the average performance of $19$ \textsc{GPTScore} based evaluators with instruction (IST) significantly outperforms vanilla (VAL). 
    (2) \textbf{The combination of instruction and demonstration (IDM) brings gains for the evaluator with different model structures.} In \autoref{tab:res-mqm}, the performance of GPT3, GPT2, OPT, and FT5  improves a lot when instruction and demonstration (IDM) are introduced. 
    (3)  \textbf{The evaluator built based on GPT3-c01 achieves comparable performance with GPT3-d01 and GPT3-d03.} This can be found in \autoref{fig:res-mqm-gpt3}.
    Since the GPT3-d01 and GPT3-d03 are most expensive variant of GPT3, the cheaper and comparative GPT3-c01 is a good choice for machine translation task.

\begin{table}[htb]
  \centering \footnotesize
   \renewcommand\tabcolsep{1pt}
    \begin{tabular}{lccccccccc}
    \toprule
    \multirow{2}[2]{*}{Model} & \multicolumn{3}{c}{ACC} & \multicolumn{3}{c}{FLU} & \multicolumn{3}{c}{MQM} \\
    \cmidrule(lr){2-4}\cmidrule(lr){5-7}\cmidrule(lr){8-10}
          & VAL & IST & IDM & VAL & IST & IDM & VAL & IST & IDM \\
    \midrule
    GPT3 & 27.2  & 27.1  & $\text{29.7}^{\dag,\ddag}$ & \textbf{11.3}  & 10.4  & $\text{\textbf{16.4}}^{\dag,\ddag}$ & 30.3  & $\text{31.2}^{\dag}$ & $\text{32.3}^{\dag,\ddag}$ \\
    GPT2 & 25.8  & $\text{27.0}^{\dag}$ & $\text{\textbf{30.3}}^{\dag,\ddag}$ & 9.8   & $\text{10.8}^{\dag}$ & $\text{15.8}^{\dag,\ddag}$ & 30.1  & $\text{30.3}^{\dag}$ & $\text{33.5}^{\dag,\ddag}$ \\
    OPT & \textbf{28.7}  & $\text{\textbf{29.4}}^{\dag}$ & $\text{30.3}^{\dag,\ddag}$ & 10.0  & $\text{\textbf{12.2}}^{\dag}$ & $\text{16.3}^{\dag,\ddag}$ & \textbf{32.5}  & $\text{\textbf{34.6}}^{\dag}$ & $\text{\textbf{35.1}}^{\dag,\ddag}$ \\
    FT5 & 27.7  & $\text{27.8}^{\dag}$ & $\text{28.3}^{\dag,\ddag}$ & 9.6   & $\text{11.0}^{\dag}$ & $\text{15.4}^{\dag,\ddag}$ & 31.0  & $\text{32.3}^{\dag}$ & 32.3 \\
    \midrule
    \rowcolor{palepink}  Avg. & 27.4  & $\text{27.8}^{\dag}$ & $\text{29.7}^{\dag,\ddag}$ & 10.2  & $\text{11.1}^{\dag}$ & $\text{16.0}^{\dag,\ddag}$ & 31.0  & $\text{32.1}^{\dag}$ & $\text{33.3}^{\dag,\ddag}$ \\
    \bottomrule
    \end{tabular}%
    \vspace{-6pt}
  \caption{The average Spearman correlation of the GPT3-based, GPT2-based, OPT-based, and FT5-based models in machine translation task of MQM-2020 dataset.
       }
  \label{tab:res-mqm}%
\end{table}%

\begin{figure}[htb]
  \centering \footnotesize
  \renewcommand\tabcolsep{0.5pt}
  \renewcommand\arraystretch{0.93}  
    \begin{tabular}{ccccccccc}
    \toprule
    \multicolumn{9}{c}{MQM-2020}  \\        
    \midrule
    \multicolumn{3}{c}{ACC} & \multicolumn{3}{c}{FLU} & \multicolumn{3}{c}{MQM} \\
    \midrule
    \multicolumn{3}{c}{\multirow{5}[2]{*}{\includegraphics[scale=0.3]{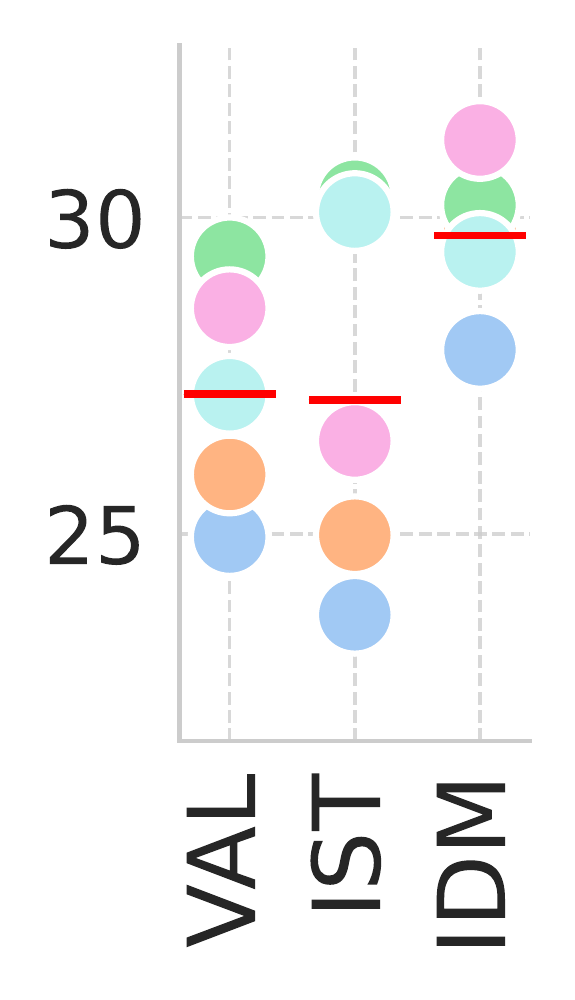}}} 
    & \multicolumn{3}{c}{\multirow{5}[2]{*}{\includegraphics[scale=0.3]{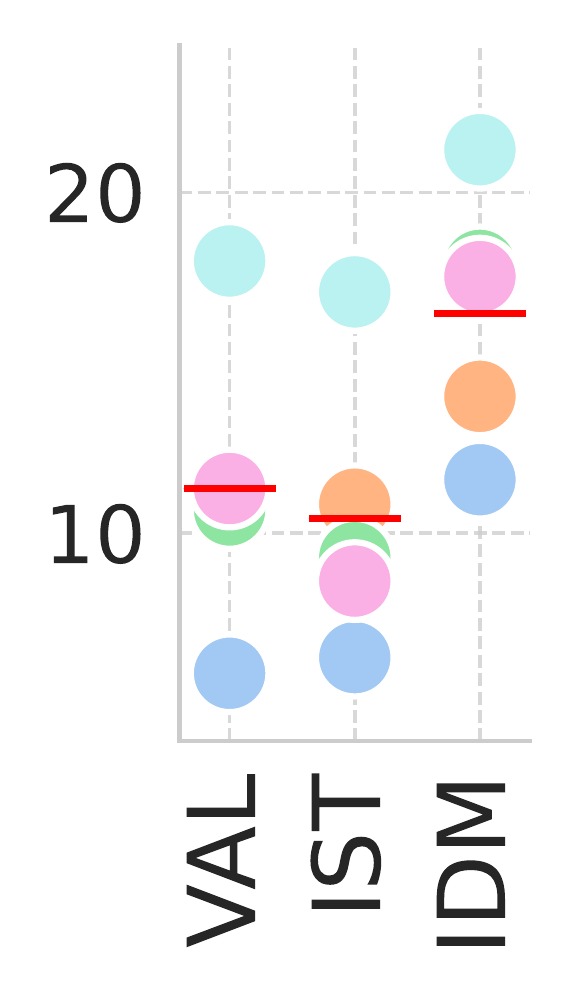}}} 
    & \multicolumn{3}{c}{\multirow{5}[2]{*}{\includegraphics[scale=0.3]{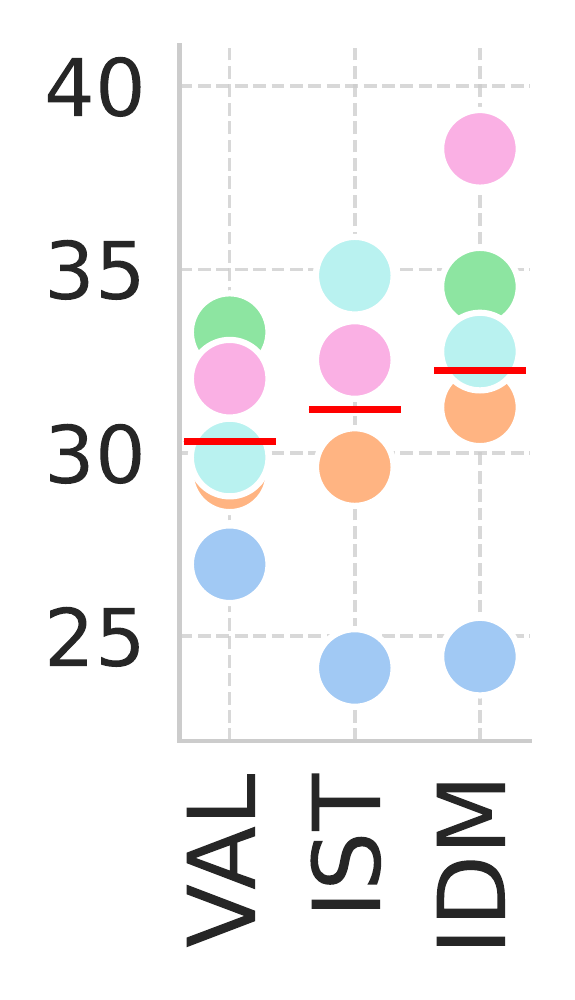}}}  \\ \\ \\ \\ \\ \\ \\ \\ \\
    \bottomrule
    \end{tabular}%
    \vspace{-6pt}
  \caption{Experimental results for GPT3-based variants in the machine translation task. Here, 
 blue,
 orange,
 green, 
 pink, 
 and 
 cyan dot denote that \textsc{GPTScore} is built based on 
 a01 (\textcolor{a01}{\faCircle}), 
 b01 (\textcolor{b01}{\faCircle}), 
 c01 (\textcolor{c01}{\faCircle}), 
 d01 (\textcolor{d01}{\faCircle}), and 
 d03 (\textcolor{d03}{\faCircle}), respectively. The
 red lines (\textcolor{red}{\textbf{---}}) denote the average performance of GPT3-based variants. 
 }
   \label{fig:res-mqm-gpt3}%
\end{figure}%

\subsection{Data to Text}
We consider the BAGEL and SFRES datasets for the evaluation of data to text task.
The average Spearman correlations of the GPT3-based, GPT2-based, OPT-based, and FT5-based models are listed in \autoref{tab:res-d2t-avg}. VAL, IST, and IDM denote the vanilla, using instruction, and using both instruction and demonstration settings, respectively.
Due to the space limitation, the detailed performance of each evaluator considered in this work can be found in \autoref{tab:d2t-bagel-full} and \autoref{tab:d2t-sfres-full}. 
The main observations are listed as follows:

    (1) \textbf{Introducing instruction (IST) can significantly improve performance, and introducing demonstration (DM) will further improve performance.}
    In \autoref{tab:res-d2t-avg}, the average performance on the three aspects is significantly improved when adapting to the instruction, and the performance of using demonstration on \texttt{NAT} and \texttt{FLU} has further significantly improved. 
    (2) \textbf{The decoder-only model is better at utilizing demonstration to achieve high performance.} In \autoref{tab:res-d2t-avg}, compare to the encoder-decoder model FT5, the performance has a more significant improvement for the decoder-only model of GPT2 and OPT on \texttt{NAT} and \texttt{FLU} aspects after introducing DM, which holds for both BAGEL and SFRES.
    (3) \textbf{GPT3 has strong compatibility with unformatted text.} Named entities of the BAGEL dataset are replaced with a special token (e.g, X and Y ). For example, ``X is a cafe restaurant'', where ``X'' denotes the name of the cafe. When introducing IST and DM (IDM), the variants of GPT3 achieve much higher average performance than GPT2, OPT, and FT5.

\begin{table}[htb]
    \centering \footnotesize
  \renewcommand\tabcolsep{1.4pt}
    \begin{tabular}{lccccccccc}
    \toprule
    \multirow{2}[2]{*}{Model} & \multicolumn{3}{c}{INF} & \multicolumn{3}{c}{NAT} & \multicolumn{3}{c}{FLU} \\
    \cmidrule(lr){2-4}\cmidrule(lr){5-7}\cmidrule(lr){8-10}
          & VAL & IST & IDM & VAL & IST & IDM & VAL & IST & IDM \\
    \midrule
    \multicolumn{10}{l}{\textbf{BAGEL}} \\
    \midrule
    GPT3  & 35.4  & $\text{38.3}^{\dag}$ & $\text{43.6}^{\dag,\ddag}$ & 21.7  & $\text{26.5}^{\dag}$ & $\text{36.9}^{\dag,\ddag}$ & 30.5  & $\text{32.9}^{\dag}$ & $\text{43.4}^{\dag,\ddag}$ \\
    GPT2  & 40.8  & $\text{43.2}^{\dag}$ & 40.2  & 31.4  & $\text{33.0}^{\dag}$ & $\text{33.5}^{\dag,\ddag}$ & 36.7  & $\text{39.3}^{\dag}$ & $\text{41.3}^{\dag,\ddag}$ \\
    OPT   & 38.7  & $\text{39.3}^{\dag}$ & 38.6  & 31.4  & 30.0  & $\text{33.7}^{\dag,\ddag}$ & 37.7  & $\text{37.1}^{\dag}$ & $\text{41.5}^{\dag,\ddag}$ \\
    FT5   & 41.5  & 41.5  & 39.1  & 26.5  & $\text{29.7}^{\dag}$ & $\text{28.6}^{\dag}$ & 38.1  & $\text{41.1}^{\dag}$ & $\text{40.3}^{\dag}$ \\
    \midrule
   \rowcolor{palepink}  \textbf{Avg.} & 39.1  & $\text{40.6}^{\dag}$ & $\text{40.3}^{\dag}$ & 27.7  & $\text{29.8}^{\dag}$ & $\text{33.2}^{\dag,\ddag}$ & 35.8  & $\text{37.6}^{\dag}$ & $\text{41.6}^{\dag,\ddag}$ \\
    \midrule
    \multicolumn{10}{l}{\textbf{SFRES}} \\
    \midrule
    GPT3  & 30.4  & 25.1  & $\text{31.5}^{\dag,\ddag}$ & 25.0  & $\text{30.4}^{\dag}$ & $\text{26.5}^{\dag}$ & 31.2  & 30.9  & 26.1 \\
    GPT2  & 22.5  & $\text{25.1}^{\dag}$ & 20.5  & 31.0  & $\text{31.9}^{\dag}$& $\text{37.0}^{\dag,\ddag}$ & 20.0  & $\text{33.1}^{\dag}$ & $\text{36.2}^{\dag,\ddag}$ \\
    OPT   & 25.2  & $\text{26.9}^{\dag}$ & 24.3  & 26.2  & $\text{30.0}^{\dag}$ & $\text{36.6}^{\dag,\ddag}$ & 21.3  & $\text{25.6}^{\dag}$ & $\text{30.6}^{\dag,\ddag}$ \\
    FT5   & 24.0  & 21.9  & 19.7  & 34.3  & $\text{34.6}^{\dag}$ & $\text{36.8}^{\dag,\ddag}$ & 22.0  & 17.8  & $\text{19.7}^{\ddag}$ \\
    \midrule
    \rowcolor{palepink}  \textbf{Avg.} & 25.5  & 24.7  & 24.0  & 29.1  & $\text{31.7}^{\dag}$ & $\text{34.2}^{\dag,\ddag}$ & 23.6  & $\text{26.8}^{\dag}$ & $\text{28.2}^{\dag,\ddag}$ \\
    \bottomrule
    \end{tabular}%
    \vspace{-6pt}
  \caption{The average of Spearman correlation the models based on GPT3, GPT2, OPT, and FT5 on BAGEL and SFRES datasets in data-to-text task.
 }
  \label{tab:res-d2t-avg}%
\end{table}%

\begin{figure}[htb]
  \centering \footnotesize
  \renewcommand\tabcolsep{-1.6pt}
  \renewcommand\arraystretch{0.98}  
    \begin{tabular}{cccccccccccccccccc}
    \toprule
    \multicolumn{9}{c}{BAGEL}                                             & \multicolumn{9}{c}{SFRES} \\
    \cmidrule(lr){1-9}\cmidrule(lr){10-18}
    \multicolumn{3}{c}{INF} & \multicolumn{3}{c}{NAT} & \multicolumn{3}{c}{FLU} & \multicolumn{3}{c}{INF} & \multicolumn{3}{c}{NAT} & \multicolumn{3}{c}{FLU} \\
    \midrule
    \multicolumn{3}{c}{\multirow{5}[2]{*}{\includegraphics[scale=0.25]{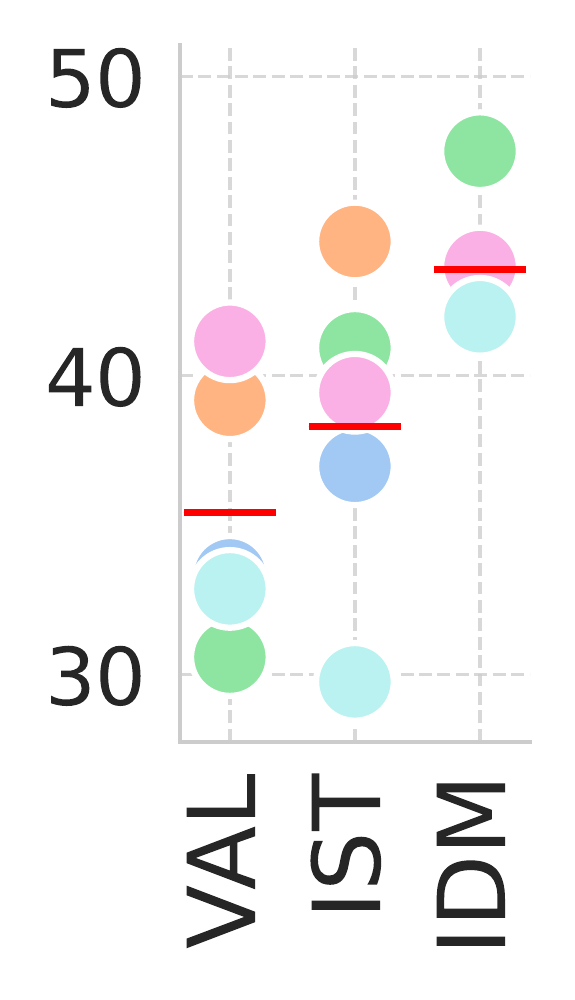}}} 
    & \multicolumn{3}{c}{\multirow{5}[2]{*}{\includegraphics[scale=0.25]{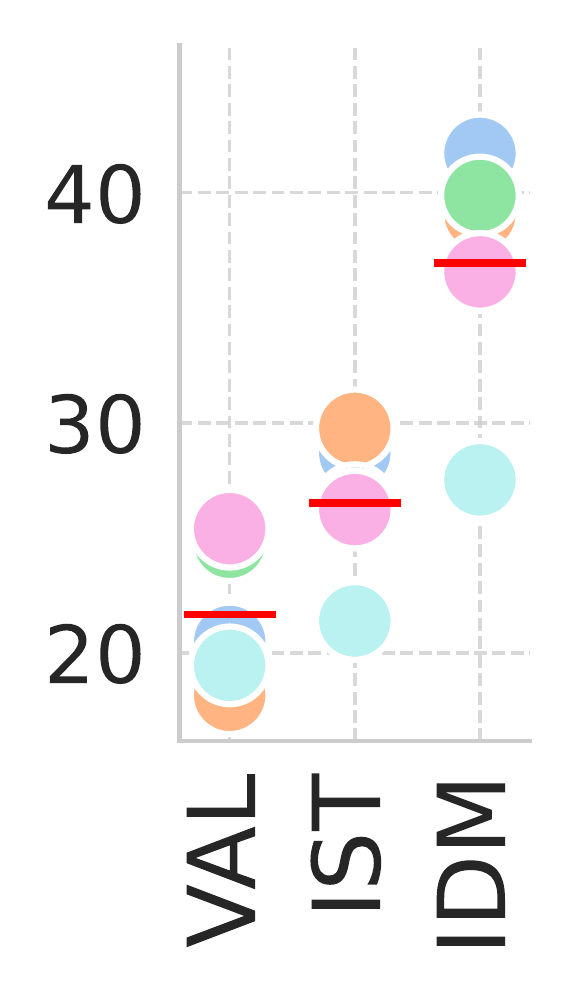}}} 
    & \multicolumn{3}{c}{\multirow{5}[2]{*}{\includegraphics[scale=0.25]{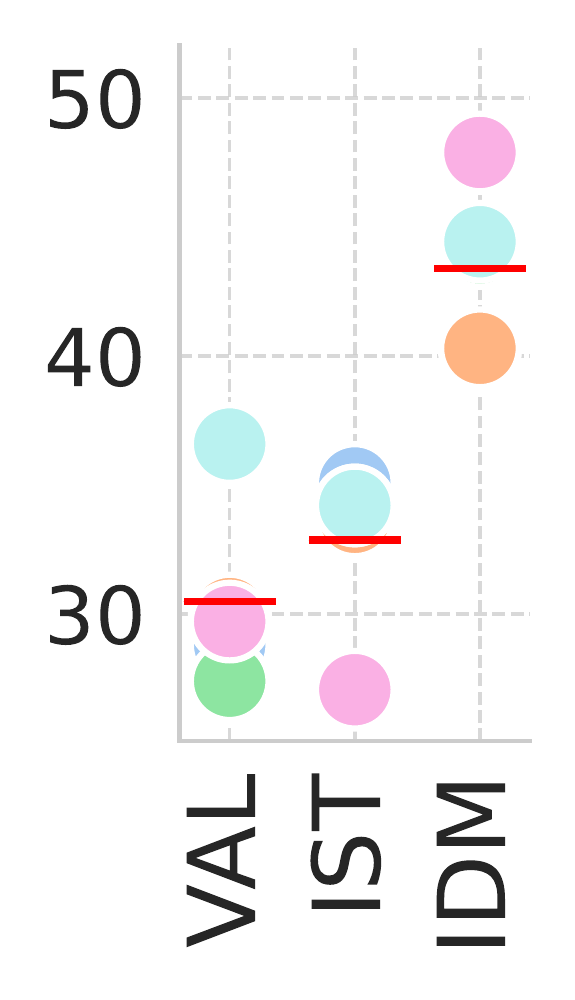}}} 
    & \multicolumn{3}{c}{\multirow{5}[2]{*}{\includegraphics[scale=0.25]{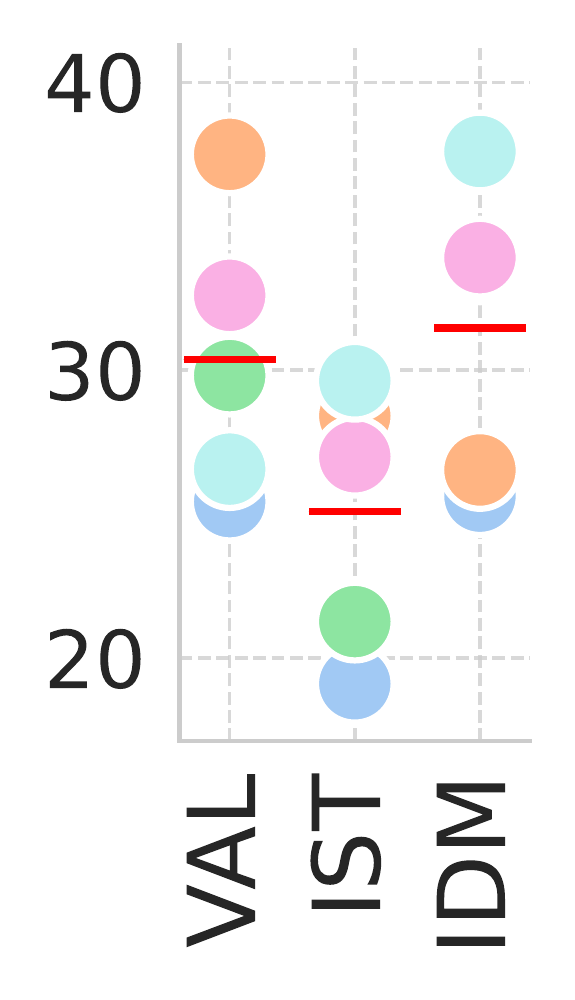}}} 
    & \multicolumn{3}{c}{\multirow{5}[2]{*}{\includegraphics[scale=0.25]{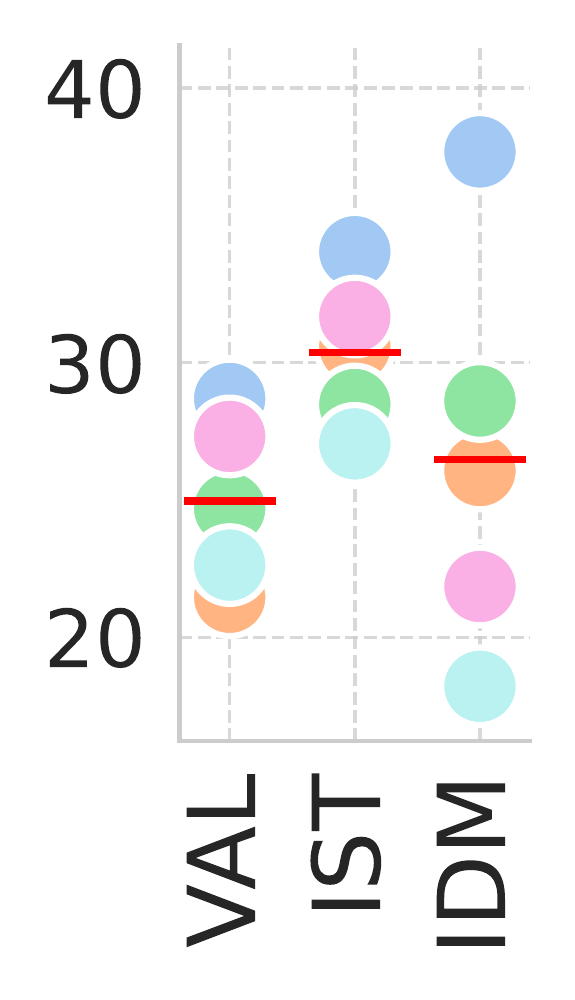}}} 
    & \multicolumn{3}{c}{\multirow{5}[2]{*}{\includegraphics[scale=0.25]{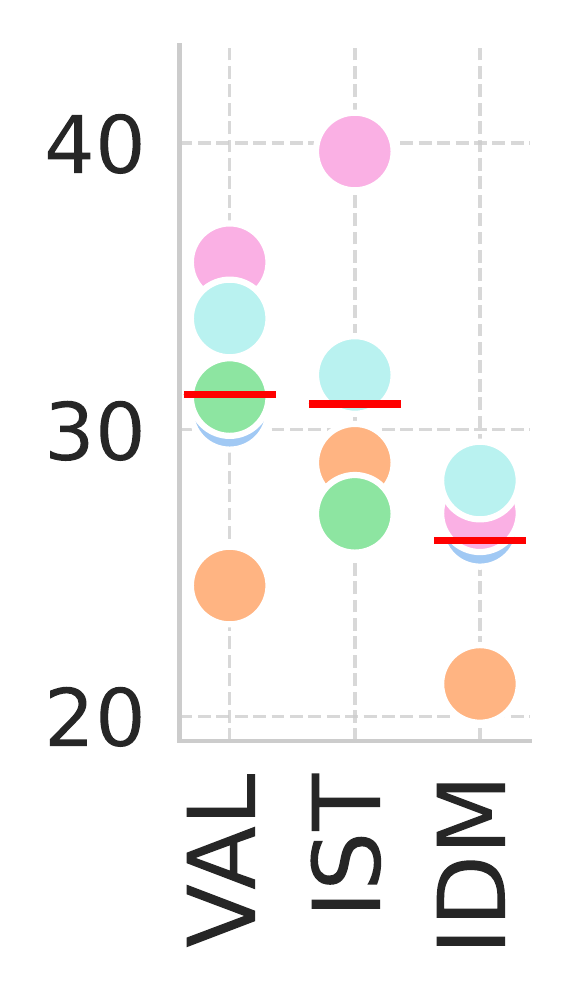}}} \\ \\ \\ \\ \\ \\ \\ 
    \bottomrule
    \end{tabular}%
    \vspace{-6pt}
  \caption{Experimental results for GPT3-based variants in data-to-text task. 
  Here, 
 blue,
 orange,
 green, 
 pink, 
 and 
 cyan dot denote that \textsc{GPTScore} is built based on 
 a01 (\textcolor{a01}{\faCircle}), 
 b01 (\textcolor{b01}{\faCircle}), 
 c01 (\textcolor{c01}{\faCircle}), 
 d01 (\textcolor{d01}{\faCircle}), and 
 d03 (\textcolor{d03}{\faCircle}), respectively. The
 red lines (\textcolor{red}{\textbf{---}}) denote the average performance of GPT3-based variants. 
 }
   \label{fig:res-d2t-gpt3}%
\end{figure}%

\subsection{Dialogue Response Generation}

To test if \textsc{GPTScore} can generalize to more aspects, we choose the task of dialogue response generation as a testbed, which usually requires evaluating generated texts from a variety of dimensions (i.e., ``interesting'' and ``fluent''). To reduce the computational cost, in this experiment, we focus on GPT3-based metrics since they have achieved superior performance as we observed in the previous experiments.

\autoref{tab:diag-spear} shows the Spearman correlation of different aspects on FED turn- and dialogue-level datasets.
The main observations are listed as follows.

    (1) \textbf{The performance of GPT3-d01 is much better than GPT3-d03, even though both of them have the same model size.} 
    The average Spearman correlation of GPT3-d01 outperforms GPT3-d03 by $\textbf{40.8}$ on the FED Turn-level dataset, and $\textbf{5.5}$ on the FED dialogue-level. 
    (2)  \textbf{The GPT3-based model demonstrate stronger generalization ability.} BART-based models failed in the evaluation of the dialogue generation task, while the GPT3-a01 with 350M parameters achieved comparable performance to FED and DE models on both the FED turn-level and dialogue-level datasets.

\begin{table}[htb]
  \centering \footnotesize
\renewcommand\tabcolsep{1.4pt}
    \begin{tabular}{lcccccccccc}
    \toprule
    \multirow{2}[2]{*}{Aspect} & \multicolumn{5}{c}{Baseline} & \multicolumn{5}{c}{GPTScore} \\
    \cmidrule(lr){2-6}\cmidrule(lr){7-11}
          & BT & BTC & BTCP & FED & DE & a01 & b01 & c01 & d01 & d03 \\
    \midrule
    \multicolumn{11}{l}{\textit{\textbf{FED dialogue-level }}} \\
    \midrule
    COH   & 1.7   & -14.9 & -18.9 & 25.7  & 43.7  & 18.7  & 15.0  & 22.5  & \textbf{56.9} & 13.4 \\
    ERR   & 9.4   & -12.2 & -13.7 & 12.0  & 30.2  & 35.2  & 16.8  & 21.3  & \textbf{45.7} & 9.40 \\
    CON   & 2.6   & -6.7  & -10.2 & 11.6  & 36.7  & \textbf{33.7} & 9.9   & 18.4  & 32.9  & 18.1 \\
    DIV   & 13.3  & -2.5  & -13.9 & 13.7  & 37.8  & 14.9  & 5.20   & 21.5  & \textbf{62.8} & -6.6 \\
    DEP   & 8.2   & -6.6  & -17.6 & 10.9  & 49.8  & 9.00   & 12.9  & 28.2  & \textbf{66.9} & 34.1 \\
    LIK   & 9.9   & -6.3  & -11.8 & 37.4  & 41.6  & 26.2  & 22.0  & 32.1  & \textbf{63.4} & 18.4 \\
    UND   & -11.5 & -17.6 & -18.2 & -0.3  & 36.5  & 31.2  & 40.0  & 40.0  & \textbf{52.4} & 19.6 \\
    FLE   & 9.3   & -10.2 & -10.3 & 24.9  & 38.3  & 32.7  & 44.9  & 34.6  & \textbf{51.5} & 7.20 \\
    INF   & 9.2   & -7.5  & -10.5 & 42.9  & 42.6  & 6.80   & 8.0   & 18.8  & \textbf{60.2} & 31.7 \\
    INQ   & 6.2   & -0.6  & -14.8 & 24.7  & 41.0  & 44.2  & 38.7  & 49.2  & \textbf{50.3} & -10.1 \\
    \midrule
    \rowcolor{palepink}  Avg.  & 5.8   & -8.5  & -14.0 & 20.4  & 39.8  & 25.3  & 21.3  & 28.6  & \textbf{54.3}  & 13.5 \\
    \midrule
    \multicolumn{11}{l}{\textit{\textbf{FED turn-level }}} \\
    \midrule
    INT   & 15.9  & -3.3  & -10.1 & 32.4  & 32.7  & 16.6  & 6.4   & 30.8  & \textbf{50.1} & 22.4 \\
    ENG   & 22.6  & 1.1   & -2.5  & 24.0  & 30.0  & 10.2  & 6.2   & 29.4  & \textbf{49.6} & 35.5 \\
    SPE   & 8.3   & -7.9  & -16.2 & 14.1  & \textbf{34.6} & 33.7  & 16.1  & 31.7  & 21.4  & 15.1 \\
    REL   & 11.9  & 10.0  & 19.4  & 19.9  & 26.3  & 8.6   & 10.3  & 23.8  & \textbf{45.2} & 38.0 \\
    COR   & 7.6   & 1.8   & 12.4  & 26.2  & 24.2  & 29.7  & 11.2  & 27.0  & \textbf{43.4} & 42.8 \\
    SEM   & 10.0  & 18.8  & 26.1  & -9.4  & 20.2  & 6.8   & 8.1   & 23.1  & \textbf{44.4} & 40.5 \\
    UND   & 12.0  & 8.1   & 4.5   & 1.3   & 20.0  & 6.6   & 14.8  & 23.4  & \textbf{36.5} & 31.1 \\
    FLU   & 14.0  & 17.2  & 28.4  & -13.4 & 17.1  & 16.5  & 5.7   & 14.0  & 16.0  & \textbf{36.7} \\
    \midrule
   \rowcolor{palepink}  Avg.  & 12.8  & 5.7   & 7.7   & 11.9  & 25.6  & 16.1  & 9.9   & 25.4  & \textbf{38.3}  & 32.8 \\
    \bottomrule
    \end{tabular}
    \vspace{-6pt}
  \caption{Spearman correlation of different aspects on the FED turn- and dialogue-level datasets. \textit{BT}, \textit{BTC}, \textit{BTCP}, and \textit{DE} denote BARTSCORE, BARTSCORE+CNN, BARTSCORE+CNN+Para, and DynaEval model, respectively. 
    Values in bold indicate the best performance.
    }
  \label{tab:diag-spear}
\end{table}

\section{Ablation Study}

\subsection{Effectiveness of Demonstration}

To investigate the relationship between the demonstration sample size (denote as K) and the evaluation performance, we choose the machine translation task and the GPT3-based variants with model sizes ranging from 350M to 175B for further study.

\begin{figure}[ht]
	\centering
	\subfigure[ACC]{
	\begin{minipage}[b]{0.16 \textwidth}
		\includegraphics[width=1\textwidth]{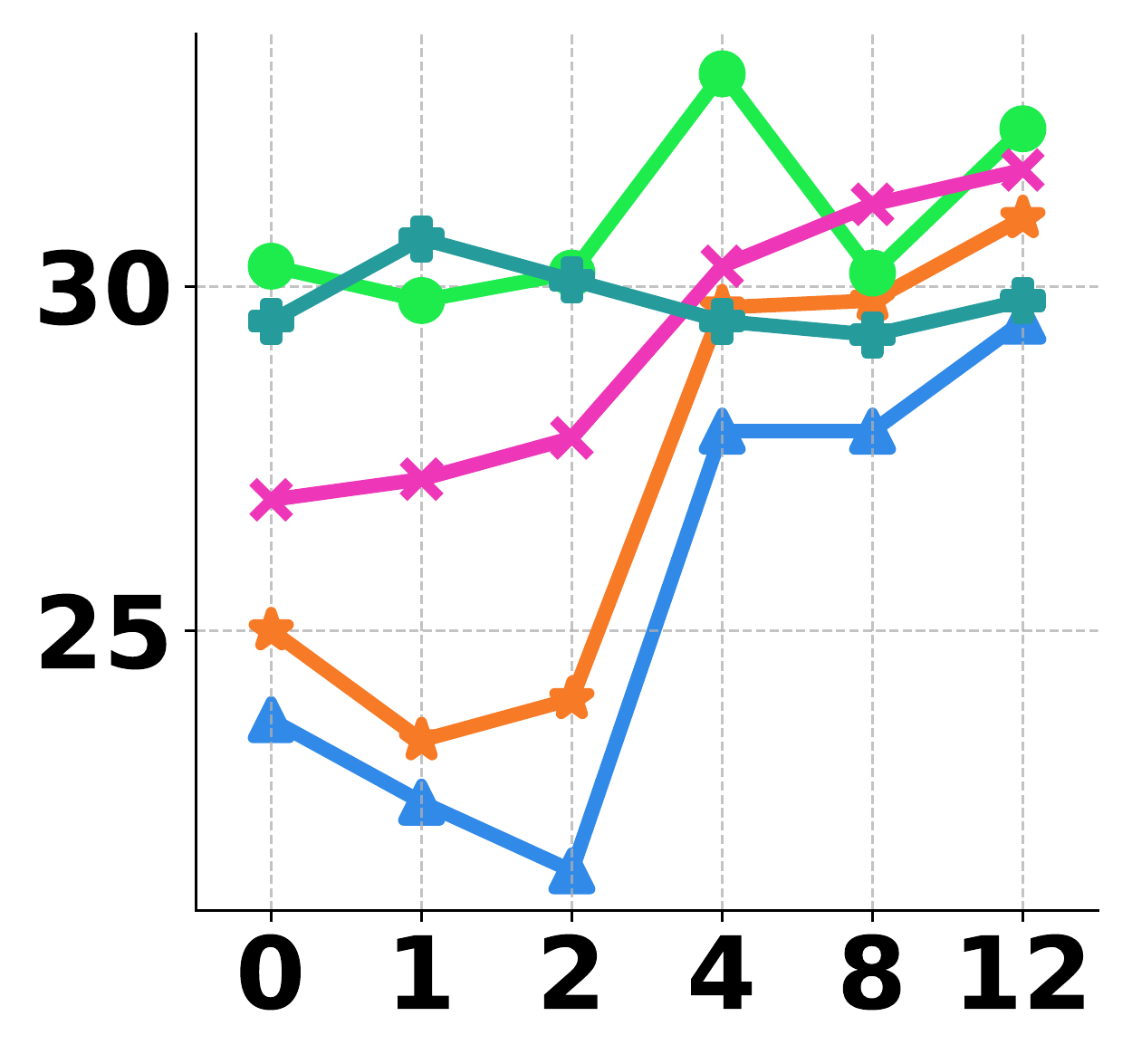}
	\end{minipage}
	} 
	\hspace{-12pt}
 \subfigure[FLU]{
	\begin{minipage}[b]{0.16 \textwidth}
		\includegraphics[width=1\textwidth]{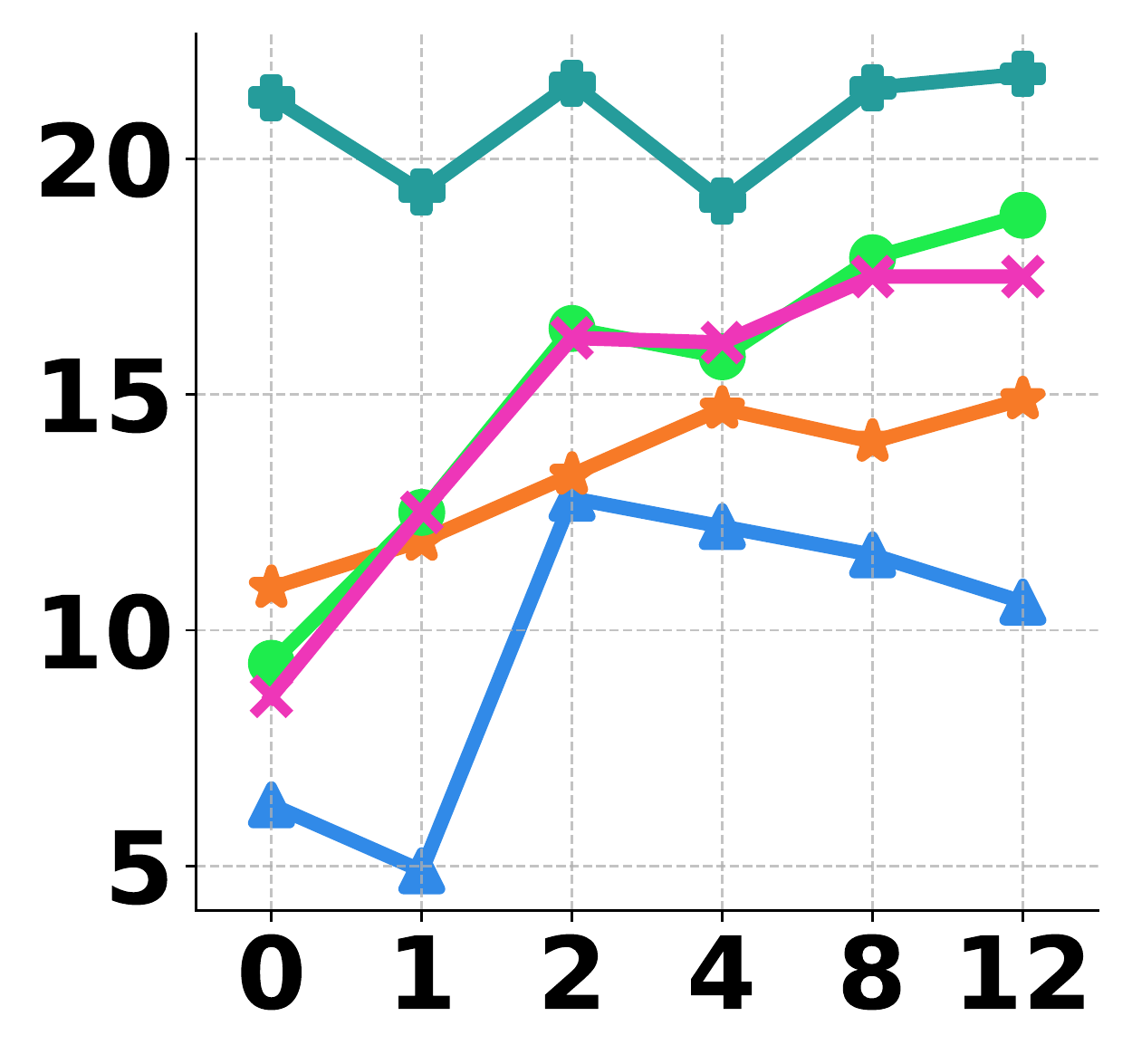}
	\end{minipage}
	}  
        \hspace{-12pt}
	\subfigure[MQM]{
	\begin{minipage}[b]{0.16 \textwidth}
		\includegraphics[width=1\textwidth]{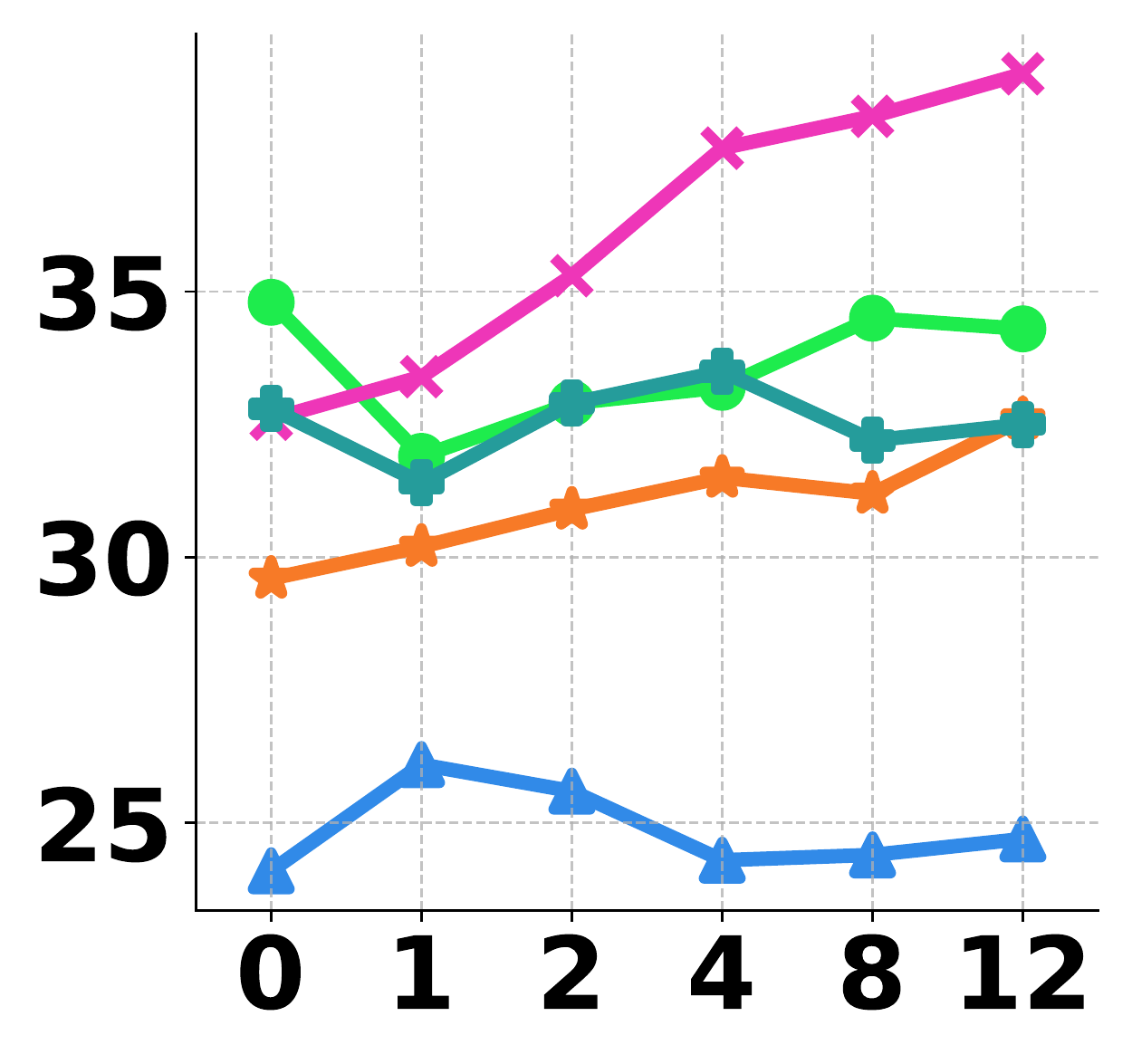}
	\end{minipage}
	}  
 \vspace{-15pt}
	\caption{Results of the GPT3 family models with different numbers of examples (K) in the demonstration on the MQM-2020 dataset. Here, 
 blue,
 orange,
 green,  
 red, and cyan lines denote that \textsc{GPTScore} is built based on GPT3-a01 (\textcolor{a01_dm}{$\blacktriangle$}), GPT3-b01 (\textcolor{b01_dm}{\faStar}), GPT3-c01 (\textcolor{c01_dm}{\faCircle}), GPT3-d01 (\textcolor{d01_dm}{\faClose}), and GPT3-d03 (\textcolor{d03_dm}{\textbf{+}}), respectively.
}
\label{fig:demon-explore}
\end{figure}

The change of Spearman correlation on the MQM-2020 dataset with different demonstration sample size are shown in \autoref{fig:demon-explore}.
The main observations are summarized as follows:
(1) The utilization of demonstration significantly improves the evaluation performance, which holds for these three aspects.
(2) There is an upper bound on the performance gains from the introduction of the demonstration. For example, when K>4, the performance of \texttt{ACC} is hard to improve further. 
(3) When DM has only a few samples (such as K=1), small models (e.g., GPT3-a01) are prone to performance degradation due to the one-sidedness of the given examples.

\subsection{Partial Order of Evaluation Aspect}
\label{sec:partial-order}
\begin{figure}[ht]
	\centering
	 \subfigure[Aspect order]{
	\begin{minipage}[b]{0.23 \textwidth}
		\includegraphics[width=1\textwidth]{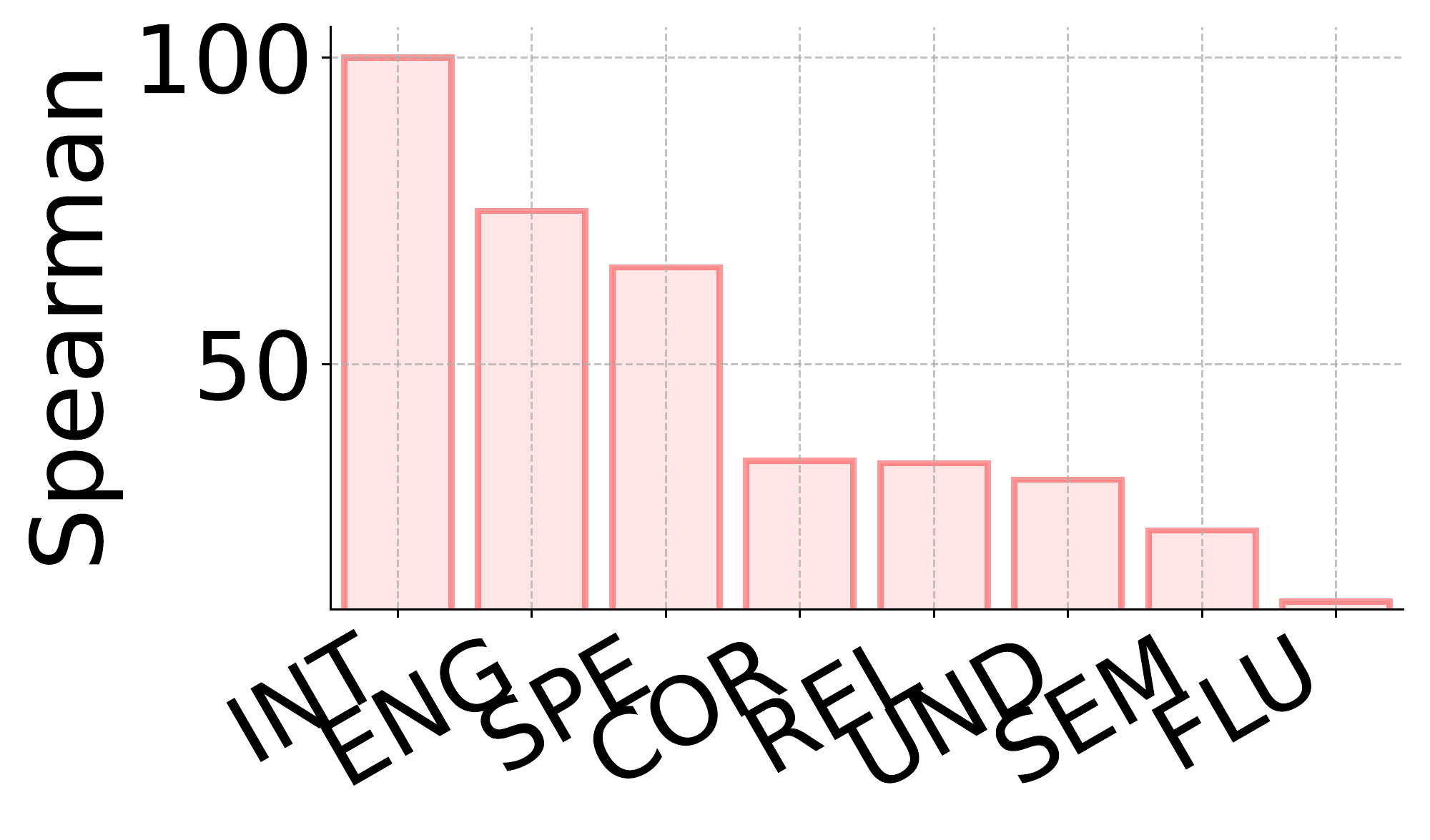}
	\end{minipage}
	}  
	\hspace{-8pt}
    \subfigure[\texttt{INT} performance]{
	\begin{minipage}[b]{0.23 \textwidth}
		\includegraphics[width=1\textwidth]{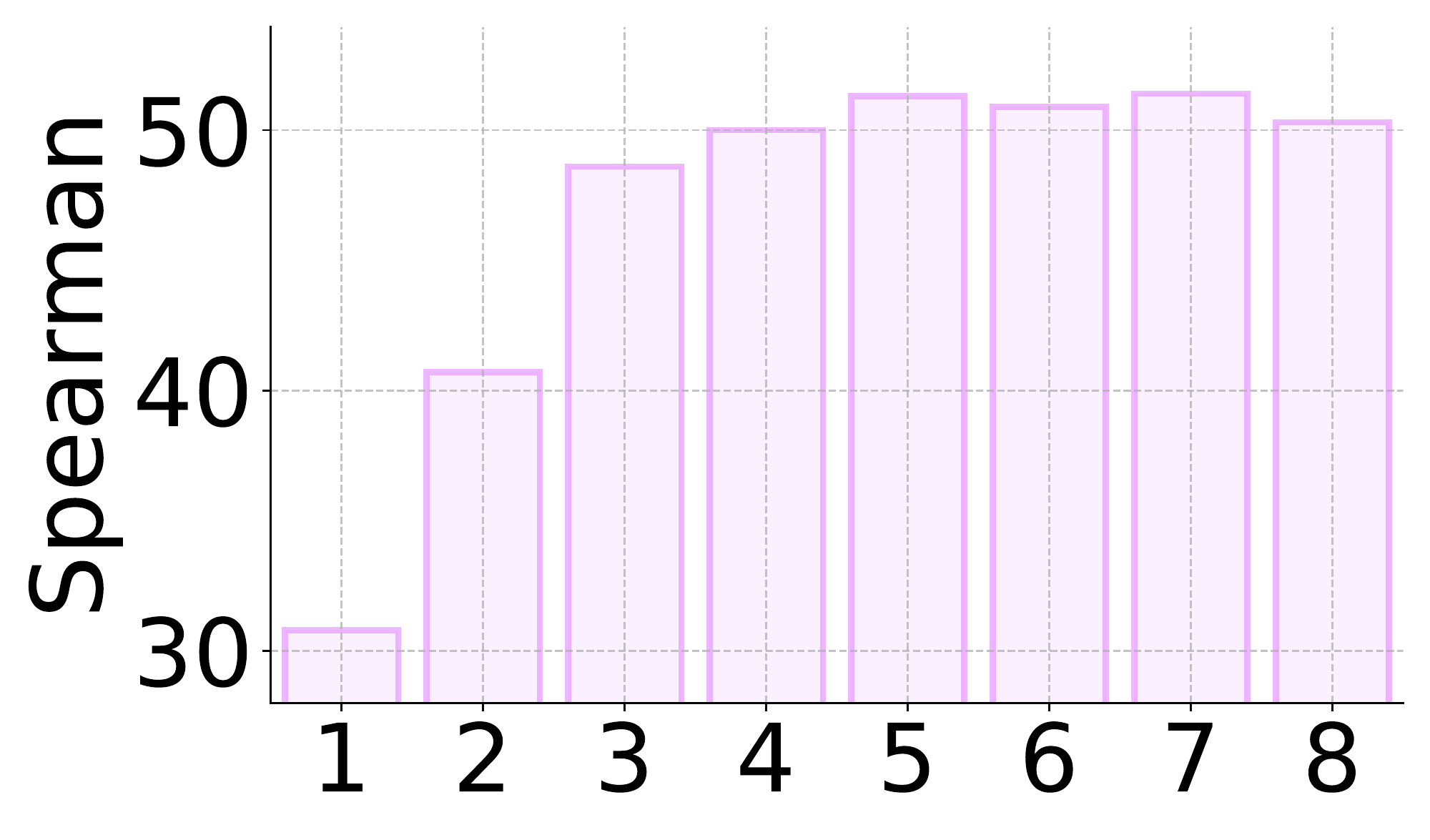}
	\end{minipage}
	} 
  \vspace{-8pt}
     \caption{(a) Descending order of Spearman correlation between \texttt{INT} and other aspects' human scoring.
(b) The Spearman correlation of \texttt{INT} changes as its aspect definition is modified in combination with other aspects. The scoring model is GPT3-c01.
     }
	\label{fig:aspect-relation}
\end{figure}

To explore the correlation between aspects, we conducted an empirical analysis with \texttt{INT} ({\textit{interesting}}) on the dialogue response generation task of the FED-Turn dataset.
Specifically, take \texttt{INT} as the target aspect and then combine the definitions of other aspects with the definition of \texttt{INT} as the final evaluation protocols.
The x-axis of \autoref{fig:aspect-relation}-(a) is the aspect order achieved based on the Spearman correlation between \texttt{INT} and that aspect's human score.
\autoref{fig:aspect-relation}-(b) is the Spearman correlation o \texttt{INT} as the modification of the \texttt{INT} definition, and the scoring function is GPT3-c01.

The following table illustrates the definition composition process, where \texttt{Sp} denotes Spearman.

{
  \centering \footnotesize
  \renewcommand\tabcolsep{2pt}
    \begin{tabular}{lp{5em}p{16em}c}
    \toprule
    \textbf{X} & \textbf{Aspect} & \textbf{Aspect Definition} & \textbf{Sp} \\
    \midrule
    1     & INT   & Is this response \texttt{interesting} to the conversation? & 30.8 \\
    \midrule
    3     & INT, ENG, SPE & Is this an \texttt{interesting} response that is \texttt{specific} and \texttt{engaging}? & 48.6 \\
    \bottomrule
    \end{tabular}
  \label{tab:addlabel}
}

Specifically, the definition of \texttt{INT} is \textit{``Is this response interesting to the conversation? ''} at x=1 in \autoref{fig:aspect-relation}-(b).
When \texttt{INT} combines with \texttt{ENG, SPE} (at x=3 in \autoref{fig:aspect-relation}-(b)), its definition can be \textit{``Is this an interesting response that is specific and engaging?''}. And the new aspect definition boosts the performance from \textbf{30.8} (at x=1 in \autoref{fig:aspect-relation}-(b)) to \textbf{48.6} (at x=3 in \autoref{fig:aspect-relation}-(b)). 
The best performance of \textbf{51.4} (x=5 in \autoref{fig:aspect-relation}-(b)) is achieved after combining five aspects (\texttt{INT, ENG, SPE, COR, REL}), which already exceeded \textbf{50.1} of the most potent scoring model GPT3-d01 with aspect definition built only on \texttt{INT}.
Therefore, combining definitions with other highly correlated aspects can improve evaluation performance.

\section{Conclusion}

In this paper, we propose to leverage the emergent abilities from generative pre-training models to address intricate and ever-changing evaluation requirements. 
The proposed framework, \textsc{GPTScore}, is studied on multiple pre-trained language models with different structures, including the GPT3 with a model size of 175B.
\textsc{GPTScore} has multiple benefits:
customizability, multi-faceted evaluation, and train-free, which enable us to flexibly craft a metric that can support 22 evaluation aspects on 37 datasets without any learning process yet attain competitive performance. This work opens a new way to audit generative AI by utilizing generative AI.

\section*{Acknowledgements}
We thank Chen Zhang for helpful discussion and feedback.
This research / project is supported by the National Research Foundation, Singapore under its Industry Alignment Fund – Pre-positioning (IAF-PP) Funding Initiative. Any opinions, findings and conclusions or recommendations expressed in this material are those of the author(s) and do not reflect the views of National Research Foundation, Singapore. Pengfei Liu is supported by a grant from the Singapore Defence Science and Technology Agency.
 
\bibliography{custom}
\bibliographystyle{icml2023}

\newpage
\appendix
\onecolumn

\section{Metric Comparison}

\autoref{tab-organization} summarize several popular generated text evaluation methods.

\begin{table*}[!ht]
  \centering \footnotesize
  \renewcommand\tabcolsep{1.7pt}
    \begin{tabular}{l@{}c@{}cccccc}
    \toprule
    \multirow{2}[0]{*}{\textbf{Metrics}} & \multirow{2}[0]{*}{\textbf{Custom}} & \multicolumn{2}{c}{\textbf{Function ($f$)}} & \multicolumn{2}{c}{\textbf{Additional text ($\mathcal{S}$)}} & \multicolumn{1}{c}{\multirow{2}[0]{*}{\textbf{Training-free}}} & \multirow{2}[0]{*}{\textbf{Application}} \\
    \cmidrule(lr){3-4}\cmidrule(lr){5-6}
          &       & \textbf{Representation} & \textbf{Formulation} & \textbf{Source} & \textbf{Reference} &       &  \\
    \midrule
    ROUGE~\cite{rouge2004lin} & \ding{55} & Token & Matching & No    & Required & \ding{51}   & SUM \\
    BLEU~\cite{bleu2002kishore}  & \ding{55}    & Token & Matching & No    & Required & \ding{51}   & MT \\
    CHRF~\cite{maja:chrf}  & \ding{55}    & Character & Matching & No    & Required & \ding{51}   & MT \\
    BERTScore~\cite{bertscore2020tianyi} & \ding{55}    & BERT  & Matching & No    & Required & \ding{51}   & MUL(2) \\
    MoverScore~\cite{moverscore2019wei} & \ding{55}    & BERT  & Matching & No    & Required & \ding{51}   & MUL(4) \\
    BLEURT~\cite{sellam:bleurt} & \ding{55}    & BERT  & Regression & No    & Required & \ding{51}   & MT \\
        \midrule
    PRISM~\cite{prism:brian} & \ding{55}  & Embedding & Paraphrase & Optional & {Optional} & \ding{51}   & MT \\
    UNIEVAL~\cite{zhong:unifiedEval} & \ding{55}  & T5    & Boolean QA & Optional & {Optional} & \ding{55}   & MUL(2) \\
    COMET~\cite{rei:comet} & \ding{55}    & BERT  & Regress, Rank & Optional & {Optional} & \ding{55}   & MT \\
    BARTScore~\cite{bartscore2021yuan} &\ding{55}  & BART  & Generation & Optional & {Optional} & \ding{51}   & MUL(3) \\
    FED~\cite{fed2020shikib}   & \ding{55}  & DialoGPT & Generation & Required & {Optional}    & \ding{51}   & Dialogue \\
    HolisticEval~\cite{pang:holisticeval} & \ding{55}  & GPT2  & Generation & Optional & {Optional}    & \ding{51}   & Dialogue \\
    \midrule
    GPTScore & \ding{51}  & GPT3/OPT  & Any & Optional & {Optional} & \ding{51}  & MUL(5) \\
    \bottomrule
    \end{tabular}
      \caption{A comprehensive comparison of existing research on automated evaluation of generated texts. MUL(k) denotes multiple (k) applications explored. \textit{Custom} denotes \textit{Custom Aspects}.
      }
      \label{tab-organization}%
\end{table*}

\section{Tasks, Datasets, and Aspects}
\label{sec:app-task-asp}

To achieve a more comprehensive evaluation, in this paper,
we cover a broad range of natural language generation tasks:
\textit{Dialogue Response Generation}, \textit{Text Summarization}, \textit{Data-to-Text}, and \textit{Machine Translation}, which involves $9$ datasets and $22$ evaluation aspects in total. \autoref{tab:task-data} summarizes the tasks, datasets, and evaluation aspects considered by each dataset. 
The definition of different aspects can be found in \autoref{tab:asp-define}.

\begin{table}[!ht]
  \centering \footnotesize
    \begin{tabular}{lll}
    \toprule
    \multicolumn{1}{l}{\textbf{Tasks}} & \textbf{Dataset} & \textbf{ Aspect} \\
    \midrule
    \multicolumn{1}{l}{\multirow{4}[4]{*}{Diag}} & \multirow{2}[2]{*}{FED-Diag} & COH, DIV, FLE, UND,INQ \\
          &       & CON, INF, LIK, DEP, ERR \\
\cmidrule(lr){2-3}          & \multirow{2}[2]{*}{FED-Turn} & INT, ENG, SPE, REL,  \\
          &       & COR, SEM, UND, FLU \\
    \midrule
    \multicolumn{1}{l}{\multirow{4}[1]{*}{Summ}} & SummEval & COH, CON, FLU,REL \\
          & Newsroom  & FLU, REL, INF, COH \\
          & REALSumm & COV \\
          & Q-XSUM & FAC \\
    \midrule
    \multicolumn{1}{l}{\multirow{2}[1]{*}{D2T}}  & BAGEL & FLU, REL, INF \\
          & SFRES  & FLU, REL, INF \\
    \midrule
    \multicolumn{1}{l}{\multirow{1}[1]{*}{MT}} & MQM-2020 & FLU, COH, INF \\
    \bottomrule
    \end{tabular}
    \caption{An overview of tasks, datasets, and evaluation aspects. \textit{Summ.} denote the text summarization task, \textit{D2T} denotes the  Data-to-Text task, \textit{MT} denotes the machine translation. \autoref{tab:asp-define} summarized the definitions of the aspects explored in this work.} 
  \label{tab:task-data}
\end{table}

\paragraph{Dialogue Response Generation} aims to automatically generate an engaging and informative response based on the dialogue history. 
(1) \texttt{FED}~\cite{fed2020shikib} collects 124 conversations, including both human-machine (Meena~\cite{meena2020daniel}, Mitsuku\footnote{\url{https://medium.com/pandorabots-blog/mitsuku-wins-loebner-prize-2018-3e8d98c5f2a7}}) and human-human dialogues, and manually annotated 9 and 11 evaluation aspects at the turn- and dialogue-level, respectively.

\paragraph{Text Summarization} is a task of automatically generating an informative and fluent summary for a given long text. Here, we consider the following four datasets covering 6 evaluation aspects: \textit{semantic coverage, informativeness, relevance, fluency, coherence}, and \textit{factuality}. 
(1) \texttt{SummEval}~\cite{realsumm2020manik} collects human judgments on 16 model-generated summaries on the CNN/Daily Mail dataset, covering aspects of coherence, consistency, fluency, and relevance.
(2) \texttt{REALSumm}~\cite{realsumm2020manik} evaluates the reliability of automatic metrics by measuring the pyramid recall of text generated by $25$ systems.
(3) \texttt{NEWSROOM}~\cite{newsroom2018max} covers news, sports, entertainment, finance, and other topics and evaluates the quality of summaries generated by 7 systems, including informativeness, relevance, fluency, and coherence.
(4) \texttt{QAGS\_XSUM}~\cite{qags20alex} is another dataset focusing on the factuality aspect. It has 239 samples from XSUM and their summaries are generated by a fine-tuned BART model.

\paragraph{Data-to-Text} aims to automatically generate a fluent and factual description for a given table. 
(1)  \texttt{BAGEL}~\cite{bagel2010francois} contains $202$ samples about restaurants in Cambridge.
(2) \texttt{SFRES}~\cite{sfres2015tsung} contains $581$ samples about restaurants in San Francisco.
These two datasets consider three evaluation aspects: \textit{informativeness}, \textit{naturalness} (relevance), and \textit{quality} (fluency).

\paragraph{Machine Translation} aims to translate a sentence from one language to another. We consider a sub-datasets of Multidimensional Quality Metrics (MQM)~\cite{freitag:mqm}, namely, MQM-2020 (Chinese->English). 
Due to limited annotations, here, we only consider three evaluation aspects: \textit{accuracy}, \textit{fluency}, and \textit{MQM} with diverse scores.

\section{Ablation Study}

\subsection{Effectiveness of Demonstration}
The in-context learning helps a lot to achieve a good performance.
However, how does the number of samples in the demonstration impact the performance? 
We conduct a case study on the five GPT3-based models explored in this work. The experimental results are shown in \autoref{fig:demon-explore}, and the specific performance values can be seen in \autoref{tab:res-demon}.

\begin{table}[htb]
  \centering \footnotesize
    \begin{tabular}{lcccc}
    \toprule
    \textbf{Model} & \textbf{K} & \textbf{ACC} & \textbf{FLU} & \textbf{MQM} \\
    \midrule
    \multirow{6}[2]{*}{GPT3-ada} & 0     & 23.7  & 6.3   & 24.1 \\
          & 1     & 22.5  & 4.9   & 26.1 \\
          & 2     & 21.5  & 12.8  & 25.6 \\
          & 4     & 27.9  & 12.2  & 24.3 \\
          & 8     & 27.9  & 11.6  & 24.4 \\
          & 12    & 29.5  & 10.6  & 24.7 \\
    \midrule
    \multirow{6}[2]{*}{GPT3-babbage} & 0     & 25.0  & 10.9  & 29.6 \\
          & 1     & 23.4  & 11.9  & 30.2 \\
          & 2     & 24.0  & 13.3  & 30.9 \\
          & 4     & 29.7  & 14.7  & 31.5 \\
          & 8     & 29.8  & 14.0  & 31.2 \\
          & 12    & 31.0  & 14.9  & 32.6 \\
    \midrule
    \multirow{6}[2]{*}{GPT3-curie} & 0     & 30.3  & 9.3   & 34.8 \\
          & 1     & 29.8  & 12.5  & 31.9 \\
          & 2     & 30.2  & 16.4  & 32.9 \\
          & 4     & 33.1  & 15.8  & 33.2 \\
          & 8     & 30.2  & 17.9  & 34.5 \\
          & 12    & 32.3  & 18.8  & 34.3 \\
    \midrule
    \multirow{6}[2]{*}{GPT3-davinci001} & 0     & 26.9  & 8.6   & 32.6 \\
          & 1     & 27.2  & 12.5  & 33.4 \\
          & 2     & 27.8  & 16.2  & 35.3 \\
          & 4     & 30.3  & 16.1  & 37.7 \\
          & 8     & 31.2  & 17.5  & 38.3 \\
          & 12    & 31.7  & 17.5  & 39.1 \\
    \midrule
    \multirow{6}[2]{*}{GPT3-davinci003} & 0     & 29.5  & 21.3  & 32.8 \\
          & 1     & 30.7  & 19.3  & 31.4 \\
          & 2     & 30.1  & 21.6  & 32.9 \\
          & 4     & 29.5  & 19.1  & 33.5 \\
          & 8     & 29.3  & 21.5  & 32.2 \\
          & 12    & 29.8  & 21.8  & 32.5 \\
    \bottomrule
    \end{tabular}%
    \caption{Spearman correlation of the GPT3-based models (e.g, text-ada-001 and text-davinci-001) with different demonstration sample numbers on the MQM-2020 dataset .K denotes the number of samples in the demonstration.}
  \label{tab:res-demon}%
\end{table}%

\subsection{Partial Order of Evaluation Aspect}

We have investigated the combination of different evaluation aspects to achieve further performance gains in \autoref{sec:partial-order}. \autoref{tab:comb-asps} summarizes the aspect definition and Spearman correlation changes for \texttt{INT}, with the introduction of other aspects.

\begin{table*}[htb]
  \centering \footnotesize
    \begin{tabular}{llp{30em}c}
    \toprule
    \textbf{X} & \textbf{Aspect} & \textbf{Aspect Definition} & \textbf{Spear} \\
    \midrule
    1     & Interesting (INT) & Is this response interesting to the convsersation? & 36.9 \\
    2     & Engaging (ENG) & Is this an interesting response that is engaging? & 40.7 \\
    3     & Specific (SPE) & Is this an interesting response that is specific and engaging? & 48.6 \\
    4     & Correct (COR) & Is this an interesting response that is engaging, specific, and correct? & 50.0 \\
    5     & Relevant (REL) & Is this an interesting response that is specific, engaging, relevant, and correct? & \textbf{51.3} \\
    6     & Understandable (UND) & Is this an interesting response that is specific, engaging, relevant, correct, and understandable? & 50.9 \\
    7     & Semantically appropriate (SEM) & Is this an interesting response that is specific, engaging, relevant, correct, understandable, and semantically appropriate? & 51.4 \\
    8     & Fluent (FLU) & Is this an interesting response that is specific, engaging, relevant, correct, understandable, semantically appropriate, and fluent? & 50.3 \\
    \bottomrule
    \end{tabular}%
    \caption{
    The aspect definition and Spearman correlation of \texttt{INT}. \textit{X} denotes the number of aspects combined with the \texttt{INT}.
    The scoring model is GPT3-c01.}
  \label{tab:comb-asps}%
\end{table*}%

\section{Prompt Design}
\label{sec:prompt}
In this work, we have studied four popular text generation tasks: text summarization, machine translation, data-to-text, and dialogue response generation. The instructions for these tasks on different evaluation aspects are summarized in \autoref{tab:ist-nlg} and \autoref{tab:ist-nlu}.
Here, we convert the dialogue response generation task as a boolean question-answering task and incorporate the aspect definition into the question of the boolean question-answering task.

\begin{table*}[ht]
  \centering \footnotesize
  \renewcommand\tabcolsep{2.4pt}
    \begin{tabular}{llp{40em}}
    \toprule
    \textbf{Aspect} & \multicolumn{1}{l}{\textbf{Function}} & \multicolumn{1}{c}{\textbf{Instruction}} \\
    \midrule
    \multicolumn{3}{l}{\textbf{\textit{Text Summarization}}} \\
    \midrule
    \multicolumn{1}{l}{\multirow{2}[1]{*}{FAC}} & src->hypo & Generate a summary with consistent facts for the following text: \{src\}\textbackslash n\textbackslash nTl;dr\{hypo\} \\
          & ref<->hypo & Rewrite the following text with consistent facts.  \{ref/hypo\} In other words, \{hypo/ref\} \\
       \cmidrule(lr){2-3} 
    \multicolumn{1}{l}{\multirow{2}[0]{*}{COV}} & src->hypo & Generate a summary with as much semantic coverage as possible for the following text: \{src\}\textbackslash n\textbackslash nTl;dr\{hypo\} \\
          & ref<->hypo & Rewrite the following text with the same semantics. \{ref/hypo\} In other words, \{hypo/ref\} \\
       \cmidrule(lr){2-3} 
    \multicolumn{1}{l}{\multirow{2}[0]{*}{CON}} & src->hypo & Generate factually consistent summary for the following text: \{src\}\textbackslash n\textbackslash nTl;dr\{hypo\} \\ 
          & ref<->hypo & Rewrite the following text with consistent facts.  \{ref/hypo\} In other words, \{hypo/ref\} \\
    \cmidrule(lr){2-3} 
    \multicolumn{1}{l}{\multirow{2}[0]{*}{INF}} & src->hypo & Generate an informative summary that captures the key points of the following text: \{src\}\textbackslash n\textbackslash nTl;dr\{hypo\} \\
          & ref<->hypo & Rewrite the following text with its core information.  \{ref/hypo\} In other words, \{hypo/ref\} \\
    \cmidrule(lr){2-3} 
    \multicolumn{1}{l}{\multirow{2}[0]{*}{COH}} & src->hypo & Generate a coherent summary for the following text: \{src\}\textbackslash n\textbackslash nTl;dr\{hypo\} \\
          & ref<->hypo & Rewrite the following text into a coherent text.  \{ref/hypo\} In other words, \{hypo/ref\} \\
    \cmidrule(lr){2-3}       
    \multicolumn{1}{l}{\multirow{2}[0]{*}{REL}} & src->hypo & Generate a relevant summary with consistent details for the following text: \{src\}\textbackslash n\textbackslash nTl;dr\{hypo\} \\
          & ref<->hypo & Rewrite the following text with consistent details.  \{ref/hypo\} In other words, \{hypo/ref\} \\
    \cmidrule(lr){2-3} 
    \multicolumn{1}{l}{\multirow{2}[1]{*}{FLU}} & src->hypo & Generate a fluent and grammatical summary for the following text: \{src\}\textbackslash n\textbackslash nTl;dr\{hypo\}\\
          & ref<->hypo & Rewrite the following text into a fluent and grammatical text. \{ref/hypo\} In other words, \{hypo/ref\} \\
    \midrule
    \midrule
    \multicolumn{3}{l}{\textbf{\textit{Machine Translation}}} \\
    \midrule
    Acc & ref<->hypo &  Rewrite the following text with its core information and consistent facts:\{ref/hypo\} In other words, \{hypo/ref\} \\
     FLU & ref<->hypo &  Rewrite the following text to make it more grammatical and well-written:\{ref/hypo\} In other words, \{hypo/ref\} \\
     MQM  & ref<->hypo &  Rewrite the following text into high-quality text with its core information:\{ref/hypo\} In other words, \{hypo/ref\} \\
    \midrule
    \midrule
    \multicolumn{3}{l}{\textbf{\textit{Data to Text}}} \\
    \midrule
    INF & ref<->hypo &  Convert the following text to another expression that preserves key information:\textbackslash n\textbackslash n\{ref/hypo\} In other words, \{hypo/ref\} \\
    NAT & ref<->hypo &  Convert the following text into another expression that is human-like and natural:\textbackslash n\textbackslash n\{ref/hypo\} In other words, \{hypo/ref\} \\
    FLU & ref<->hypo &  Convert the following text into another expression that preserves key information and is human-like and natural:\textbackslash n\textbackslash n\{ref/hypo\} In other words, \{hypo/ref\} \\
    \bottomrule
    \end{tabular}%
    \caption{Instruction design on different aspects for text summarization, machine translation, and data-to-text tasks. \textit{src}, \textit{hypo}, and \textit{ref} denote the \textit{source text}, \textit{hypothesis text}, and \textit{reference text}, respectively. \textit{a->b} (\textit{a<-b}) denotes to evaluate the quality of \textit{b} (\textit{a}) text based on the given \textit{a} (\textit{b}) text.}
  \label{tab:ist-nlg}%
\end{table*}%

\begin{table*}[ht] 
  \centering \footnotesize
    \renewcommand\tabcolsep{2.4pt}
    \begin{tabular}{lp{42em}}
    \toprule
    \textbf{Aspect} & \textbf{Instruction} \\
    \midrule
    \multicolumn{2}{l}{\textbf{\textit{FED Turn-Level}}} \\
    \midrule
    \multicolumn{1}{l}{\multirow{2}[1]{*}{INT}}  & Answer the question based on the conversation between a human and AI.\textbackslash nQuestion: Are the responses of AI interesting? (a) Yes. (b) No.\textbackslash nConversation: \{History\}\textbackslash nAnswer: Yes. \\
    \cmidrule(lr){2-2}
    \multicolumn{1}{l}{\multirow{2}[1]{*}{ENG}}   & Answer the question based on the conversation between a human and AI.\textbackslash nQuestion: Are the responses of AI engaging? (a) Yes. (b) No.\textbackslash nConversation: \{History\}\textbackslash nAnswer: Yes. \\
    \cmidrule(lr){2-2}
    \multicolumn{1}{l}{\multirow{2}[1]{*}{UND}}   & Answer the question based on the conversation between a human and AI.\textbackslash nQuestion: Are the responses of AI understandable? (a) Yes. (b) No.\textbackslash nConversation: \{History\}\textbackslash nAnswer: Yes. \\
    \cmidrule(lr){2-2}
    REL & Answer the question based on the conversation between a human and AI.\textbackslash nQuestion: Are the responses of AI relevant to the conversation? (a) Yes. (b) No.\text backslash{}nConversation: \{History\}\textbackslash nAnswer: Yes. \\
    \cmidrule(lr){2-2}
    SPE & Answer the question based on the conversation between a human and AI.\textbackslash nQuestion: Are the responses of AI generic or specific to the conversation? (a) Yes. (b) No.\textbackslash nConversation: \{History\}\textbackslash nAnswer: Yes. \\
    \cmidrule(lr){2-2}
    COR & Answer the question based on the conversation between a human and AI.\textbackslash nQuestion: Are the responses of AI correct to conversations? (a) Yes. (b) No.\textbackslash nConversation: \{History\}\textbackslash nAnswer: Yes.] \\
    \cmidrule(lr){2-2}
    SEM & Answer the question based on the conversation between a human and AI.\textbackslash nQuestion: Are the responses of AI semantically appropriate? (a) Yes. (b) No.\textbackslash nConversation: \{History\}\textbackslash nAnswer: Yes. \\
    \cmidrule(lr){2-2}
    FLU & Answer the question based on the conversation between a human and AI.\textbackslash nQuestion: Are the responses of AI fluently written? (a) Yes. (b) No.\textbackslash nConversation: \{History\}\textbackslash nAnswer: Yes. \\
    \midrule
    \midrule
    \multicolumn{2}{l}{\textbf{\textit{FED Dialog-Level}}} \\
    \midrule
    COH & Answer the question based on the conversation between a human and AI.\textbackslash nQuestion: Is the AI coherent and maintains a good conversation flow throughout the conversation? (a) Yes. (b) No.\textbackslash nConversation: \{History\}\textbackslash nAnswer: Yes. \\
    \cmidrule(lr){2-2}
    DIV & Answer the question based on the conversation between a human and AI.\textbackslash nQuestion: Is there diversity in the AI responses? (a) Yes. (b) No.\textbackslash nConversation: \{History\}\textbackslash nAnswer: Yes. \\
    \cmidrule(lr){2-2}
    FLE & Answer the question based on the conversation between a human and AI.\textbackslash nQuestion: Is the AI flexible and adaptable to human and their interests? (a) Yes. (b) No. \textbackslash nConversation: \{History\}\textbackslash nAnswer: Yes. \\
    \cmidrule(lr){2-2}
    UND & Answer the question based on the conversation between a human and AI.\textbackslash nQuestion: Does the AI seem to understand the human? (a) Yes. (b) No. \textbackslash nConversation: \{History\}\textbackslash nAnswer: Yes. \\
    \cmidrule(lr){2-2}
    INQ & Answer the question based on the conversation between a human and AI.\textbackslash nQuestion: Is the AI inquisitive throughout the conversation? (a) Yes. (b) No.\textbackslash nConversation: \{History\}\textbackslash nAnswer: Yes. \\
    \cmidrule(lr){2-2}
    CON & Answer the question based on the conversation between a human and AI.\textbackslash nQuestion: Are the responses of AI consistent in the information it provides throughout the conversation? (a) Yes. (b) No.\textbackslash nConversation: \{History\}\textbackslash nAnswer: Yes. \\
    \cmidrule(lr){2-2}
    INF & nswer the question based on the conversation between a human and AI.\textbackslash nQuestion: Are the responses of AI informative throughout the conversation? (a) Yes. (b) No.\textbackslash nConversation: \{History\}\textbackslash nAnswer: Yes. \\
    \cmidrule(lr){2-2}
    LIK & Answer the question based on the conversation between a human and AI.\textbackslash nQuestion: Does the AI display a likeable personality? (a) Yes. (b) No.\textbackslash nConversation: \{History\}\textbackslash nAnswer: Yes. \\
    \cmidrule(lr){2-2}
    DEP & Answer the question based on the conversation between a human and AI.\textbackslash nQuestion: Does the AI discuss topics in depth? (a) Yes. (b) No.\textbackslash nConversation: \{History\}\textbackslash nAnswer: Yes. \\
    \cmidrule(lr){2-2}
    ERR & Answer the question based on the conversation between a human and AI.\textbackslash nQuestion: Is the AI able to recover from errors that it makes? (a) Yes. (b) No.\textbackslash nConversation: \{History\}\textbackslash nAnswer: Yes. \\
    \bottomrule
    \end{tabular}
    \caption{Instruction design on various aspects for dialogue response generation task at the turn- and dialogue-level. \textit{History} indicates the conversation history. We convert the evaluation of the response generation task as a question-answering task, and the aspect definition is incorporated into the question of the question-answering task.
    }
  \label{tab:ist-nlu}
\end{table*}

\section{Experiment Results}
\label{sec:exp-res-full}

This section lists the full experimental results for the explored text generation tasks.
The models considered here include the 9 baseline models: ROUGE-1, ROUGE-2, ROUGE-L, BERTScore, MoverScore, PRISM, BARTSCORE, BARTSCORE+CNN, and BARTSCORE+CNN+Para,
and $19$ GPTScore models built based on the GPT3-based, GPT2-based, OPT-based, and FLAN-T5-based pre-trained models.

\autoref{tab:summ-news-full} lists the results of the text summarization datasets.
\autoref{tab:mqm2020-full} lists the results of the machine translation datasets.
\autoref{tab:d2t-bagel-full} shows the results of the data-to-text task on the BAGEL dataset.
\autoref{tab:d2t-sfres-full} shows the results of the data-to-text task on the SFRES dataset.

\begin{table*}[htb]
  \centering \footnotesize 
    \begin{tabular}{lcccccccccc}
    \toprule
    \multirow{3}[4]{*}{\textbf{Model}} & \multicolumn{8}{c}{\textbf{NEWSROOM}}                         & \multicolumn{2}{c}{\textbf{QXSUM}} \\
\cmidrule(lr){2-9} \cmidrule(lr){10-11}            & \multicolumn{2}{c}{\textbf{COH}} & \multicolumn{2}{c}{\textbf{CON}} & \multicolumn{2}{c}{\textbf{FLU}} & \multicolumn{2}{c}{\textbf{REL}} & \multicolumn{2}{c}{\textbf{COV}} \\
          & \textbf{VAL} & \textbf{IST} & \textbf{VAL} & \textbf{IST} & \textbf{VAL} & \textbf{IST} & \textbf{VAL} & \textbf{IST} & \textbf{VAL} & \textbf{IST} \\
    \midrule
    ROUGE-1 & 27.3  & -     & 26.1  & -     & 25.9  & -     & 34.4  & -     & 3.6   & - \\
    ROUGE-2 & 10.9  & -     & 11.7  & -     & 11.2  & -     & 14.4  & -     & 9.9   & - \\
    ROUGE-L & 24.7  & -     & 25.7  & -     & 24.4  & -     & 32.5  & -     & 5.2   & - \\
    BERTScore & 31.7  & -     & 31.7  & -     & 27.2  & -     & 33.7  & -     & -4.6  & - \\
    MoverScore & 17.7  & -     & 14.2  & -     & 16.0  & -     & 18.9  & -     & 5.4   & - \\
    PRISM & 60.7  & -     & 56.5  & -     & 59.2  & -     & 61.9  & -     & 2.5   & - \\
    BARTSCORE & 70.3  & -     & 67.2  & -     & 63.1  & -     & 68.8  & -     & 0.9   & - \\
    +CNN  & 68.5  & -     & 64.9  & -     & 60.4  & -     & 66.3  & -     & 18.4  & - \\
    +CNN+Para & 69.0  & -     & 65.5  & -     & 62.5  & -     & 67.3  & -     & 6.4   & - \\
    \midrule
    \textbf{GPT3} &       &       &       &       &       &       &       &       &       &  \\
    \midrule
    GPT3-a01 & 71.6  & $\text{71.9}^{\dag}$  & 69.7  & $\text{70.0}^{\dag}$  & 66.0  & $\text{67.0}^{\dag}$  & 69.6  & 69.2  & 10.3  & 9.2 \\
    GPT3-b01 & 73.6  & 72.9  & 70.2  & 70.3  & 66.8  & $\text{68.3}^{\dag}$  & 71.5  & 71.2  & 8.5   & 14.2 \\
    GPT3-c01 & \textbf{73.8} & 72.8  & \textbf{70.5} & $\text{\textbf{70.9}}^{\dag}$  & 65.9  & $\text{68.6}^{\dag}$  & 71.0  & 71.1  & 15.2  & $\text{22.1}^{\dag}$  \\
    GPT3-d01 & 72.6  & $\text{\textbf{73.4}}^{\dag}$  & 68.5  & $\text{70.0}^{\dag}$  & 65.9  & $\text{66.9}^{\dag}$  & 71.1  & $\text{72.1}^{\dag}$  & \textbf{24.0} & \textbf{22.7} \\
    GPT3-d03 & 73.8  & 73.1  & 70.4  & 70.0  & \textbf{67.4} & $\text{\textbf{68.9}}^{\dag}$  & \textbf{74.1} & \textbf{73.3} & 21.7  & $\text{22.0}^{\dag}$  \\
    \midrule
    \rowcolor{palepink} \textbf{Avg.} & 73.1  & 72.8  & 69.9  & $\text{70.2}^{\dag}$  & 66.4  & $\text{67.9}^{\dag}$  & 71.4  & 71.4  & 15.9  & $\text{18.0}^{\dag}$  \\
    \midrule
    \textbf{GPT2} &       &       &       &       &       &       &       &       &       &  \\
    \midrule
    GPT2-M & 68.9  & $\text{71.7}^{\dag}$  & 66.4  & $\text{68.0}^{\dag}$  & 61.1  & $\text{62.3}^{\dag}$  & 67.0  & 66.8  & 18.1  & $\text{18.7}^{\dag}$  \\
    GPT2-L & 70.5  & $\text{\textbf{72.3}}^{\dag}$  & 66.6  & $\text{68.3}^{\dag}$  & 60.2  & $\text{61.4}^{\dag}$  & 66.8  & $\text{67.8}^{\dag}$  & 19.2  & $\text{19.6}^{\dag}$  \\
    GPT2-XL & 71.0  & 70.5  & 66.6  & 66.6  & 61.4  & 60.7  & 67.2  & 66.9  & 21.2  & 21.2 \\
    GPT-J-6B & \textbf{71.8} & 71.4  & \textbf{69.8} & \textbf{69.5} & \textbf{65.5} & \textbf{65.5} & \textbf{69.4} & \textbf{69.3} & \textbf{21.6} & $\text{\textbf{22.0}}^{\dag}$  \\
    \midrule
    \rowcolor{palepink} \textbf{Avg.} & 70.5  & $\text{71.5}^{\dag}$  & 67.4  & $\text{68.1}^{\dag}$  & 62.0  & $\text{62.5}^{\dag}$  & 67.6  & 67.7  & 20.0  & $\text{20.4}^{\dag}$  \\
    \midrule
    \textbf{OPT} &       &       &       &       &       &       &       &       &       &  \\
    \midrule
    OPT-350M & 70.6  & $\text{71.5}^{\dag}$  & 69.2  & $\text{69.9}^{\dag}$  & 67.3  & $\text{68.1}^{\dag}$  & 70.8  & $\text{71.6}^{\dag}$  & 13.5  & 13.3 \\
    OPT-1.3B & \textbf{73.2} & $\text{\textbf{73.6}}^{\dag}$  & \textbf{70.9} & $\text{\textbf{71.3}}^{\dag}$  & 67.2  & $\text{\textbf{67.8}}^{\dag}$  & \textbf{72.5} & \textbf{72.4} & 21.1  & 19.9 \\
    OPT-6.7B & 71.9  & 71.9  & 69.0  & 69.0  & \textbf{67.7} & 67.1  & 71.7  & 71.3  & 21.2  & 19.9 \\
    OPT-13B & 71.9  & 71.9  & 68.9  & $\text{69.6}^{\dag}$  & 65.4  & $\text{66.0}^{\dag}$  & 71.2  & $\text{71.5}^{\dag}$  & 23.1  & 22.1 \\
    OPT-66B & 72.8  & 72.8  & 70.0  & 69.5  & 66.0  & 65.9  & 71.9  & 71.9  & \textbf{24.0} & \textbf{23.1} \\
    \midrule
    \rowcolor{palepink} \textbf{Avg.} & 72.1  & $\text{72.3}^{\dag}$  & 69.6  & $\text{69.9}^{\dag}$  & 66.7  & $\text{67.0}^{\dag}$  & 71.6  & $\text{71.8}^{\dag}$  & 20.6  & 19.6 \\
    \midrule
    \textbf{FLAN-T5} &       &       &       &       &       &       &       &       &       &  \\
    \midrule
    FT5-S & 68.3  & $\text{69.2}^{\dag}$  & 64.6  & 64.1  & 59.8  & $\text{60.4}^{\dag}$  & 64.6  & $\text{65.5}^{\dag}$  & 14.4  & $\text{15.1}^{\dag}$  \\
    FT5-B & 68.9  & 69.0  & 64.8  & 64.6  & 59.6  & $\text{59.9}^{\dag}$  & 66.5  & 66.5  & 13.6  & $\text{16.3}^{\dag}$  \\
    FT5-L & 70.5  & 69.1  & 66.1  & 64.6  & 60.9  & 60.0  & 66.6  & 65.4  & \textbf{27.2} & $\text{\textbf{28.8}}^{\dag}$  \\
    FT5-XL & \textbf{72.1} & \textbf{70.1} & \textbf{66.7} & \textbf{65.6} & \textbf{61.0} & \textbf{60.5} & \textbf{68.3} & 67.5  & 18.9  & $\text{25.6}^{\dag}$  \\
    FT5-XXL & 70.7  & 69.3  & 65.7  & 65.2  & 60.2  & $\text{60.4}^{\dag}$  & 67.6  & $\text{\textbf{67.8}}^{\dag}$  & 23.9  & $\text{27.8}^{\dag}$  \\
    \midrule
    \rowcolor{palepink} \textbf{Avg.} & 70.1  & 69.3  & 65.6  & 64.8  & 60.3  & 60.2  & 66.7  & 66.5  & 19.6  & $\text{22.7}^{\dag}$  \\
    \midrule
    \textbf{Overall Avg} & 71.5  & 71.5  & 68.1  & 68.3  & 64.0  & $\text{64.5}^{\dag}$  & 69.4  & 69.4  & 19.0  & $\text{20.2}^{\dag}$ \\
    \bottomrule
    \end{tabular}
    \caption{Spearman correlations on NEWSROOM and QXSUM datasets for text summarization task.
    VAL and IST denote the evaluator with vanilla and instruction, respectively.
    Values with $\dag$ denote the evaluator with instruction significantly outperforms with vanilla. Values in bold are the best performance in a set of variants (e.g., GPT3 family). 
    }
  \label{tab:summ-news-full}
\end{table*}

\begin{table*}[htb]
  \centering \footnotesize
    \begin{tabular}{lccccccccc} 
    \toprule
    \multirow{2}[2]{*}{\textbf{Model}} & \multicolumn{3}{c}{\textbf{ACC}} & \multicolumn{3}{c}{\textbf{FLU}} & \multicolumn{3}{c}{\textbf{MQM}} \\
    \cmidrule(lr){2-4} \cmidrule(lr){5-7} \cmidrule(lr){8-10} 
          & \textbf{VAL} & \textbf{IST} & \textbf{IDM} & \textbf{VAL} & \textbf{IST} & \textbf{IDM} & \textbf{VAL} & \textbf{IST} & \textbf{IDM} \\
    \midrule
    ROUGE-1 & 21.3  & -     & -     & 1.7   & -     & -     & 17.5  & -     & - \\
    ROUGE-2 & 15.0  & -     & -     & 5.8   & -     & -     & 15.4  & -     & - \\
    ROUGE-L & 16.6  & -     & -     & 8.7   & -     & -     & 15.7  & -     & - \\
    BERTScore & 26.1  & -     & -     & 8.2   & -     & -     & 23.6  & -     & - \\
    MoverScore & 18.2  & -     & -     & 1.2   & -     & -     & 17.2  & -     & - \\
    PRISM & 25.9  & -     & -     & 9.1   & -     & -     & 27.4  & -     & - \\
    BARTSCORE & 26.1  & -     & -     & 8.2   & -     & -     & 23.6  & -     & - \\
    +CNN  & 26.2  & -     & -     & 8.1   & -     & -     & 28.7  & -     & - \\
    +CNN+Para & 31.0  & -     & -     & 10.8  & -     & -     & 29.9  & -     & - \\
    \midrule
    \textbf{GPT3} &       &       &       &       &       &       &       &       &  \\
    \midrule
    GPT3-a01 & 24.9  & 23.7  & $\text{27.9}^{\dag,\ddag}$ & 5.9   & $\text{6.3}^{\dag}$  & $\text{11.6}^{\dag,\ddag}$ & 27.0  & 24.1  & $\text{24.4}^{\ddag}$ \\
    GPT3-b01 & 25.9  & 25.0  & $\text{29.8}^{\dag,\ddag}$ & 10.7  & 10.8  & $\text{14.0}^{\dag,\ddag}$ & 29.4  & 29.6  & $\text{31.2}^{\dag,\ddag}$ \\
    GPT3-c01 & \textbf{29.4}  & $\text{\textbf{30.3}}^{\dag}$  & $\text{30.2}^{\dag}$  & 10.7  & 9.3   & $\text{17.9}^{\dag,\ddag}$ & \textbf{33.3}  & $\text{34.8}^{\dag}$  & $\text{34.5}^{\dag}$  \\
    GPT3-d01 & 28.6  & 26.5  & $\text{\textbf{31.2}}^{\dag,\ddag}$ & 11.3  & 8.6   & $\text{17.5}^{\dag,\ddag}$ & 32.0  & $\text{32.5}^{\dag}$  & $\text{\textbf{38.3}}^{\dag,\ddag}$ \\
    GPT3-d03 & 27.2  & $\text{30.1}^{\dag}$  & $\text{29.5}^{\dag}$  & \textbf{18.0}  & \textbf{17.1}  & $\text{\textbf{21.3}}^{\dag,\ddag}$ & 29.9  & $\text{\textbf{34.8}}^{\dag}$  & $\text{32.8}^{\dag}$  \\
    \midrule
     \rowcolor{palepink} \textbf{Avg.} & 27.2  & 27.1  & $\text{29.7}^{\dag,\ddag}$ & 11.3  & 10.4  & $\text{16.4}^{\dag,\ddag}$ & 30.3  & $\text{31.2}^{\dag}$  & $\text{32.3}^{\dag,\ddag}$ \\
    \midrule
    \textbf{GPT2} &       &       &       &       &       &       &       &       &  \\
    \midrule
    GPT2-M & 25.7  & 24.6  & $\text{29.6}^{\dag,\ddag}$ & 8.6   & $\text{9.4}^{\dag}$  & $\text{15.1}^{\dag,\ddag}$ & 32.1  & 29.4  & $\text{34.1}^{\dag,\ddag}$ \\
    GPT2-L & 27.2  & $\text{28.5}^{\dag}$  & $\text{32.2}^{\dag,\ddag}$ & 11.1  & 10.4  & $\text{14.9}^{\dag,\ddag}$ & 31.2  & 30.9  & $\text{33.9}^{\dag,\ddag}$ \\
    GPT2-XL & 24.2  & $\text{27.6}^{\dag}$  & $\text{29.7}^{\dag,\ddag}$ & 9.4   & $\text{12.0}^{\dag}$  & $\text{17.4}^{\dag,\ddag}$ & 28.6  & $\text{32.2}^{\dag}$  & $\text{35.8}^{\dag,\ddag}$ \\
    GPT-J-6B & 26.2  & $\text{27.2}^{\dag}$  & $\text{29.5}^{\dag,\ddag}$ & 9.9   & $\text{11.2}^{\dag}$  & $\text{15.9}^{\dag,\ddag}$ & 28.5  & $\text{28.8}^{\dag}$  & $\text{30.3}^{\dag,\ddag}$ \\
    \midrule
     \rowcolor{palepink} \textbf{Avg.} & 25.8  & $\text{27.0}^{\dag}$  & $\text{30.3}^{\dag,\ddag}$ & 9.8   & $\text{10.8}^{\dag}$  & $\text{15.8}^{\dag,\ddag}$ & 30.1  & $\text{30.3}^{\dag}$  & $\text{33.5}^{\dag,\ddag}$ \\
    \midrule
    \textbf{OPT} &       &       &       &       &       &       &       &       &  \\
    \midrule
    OPT-350M & 29.3  & 28.1  & $\text{28.6}^{\ddag}$ & 11.7  & 11.9  & $\text{15.7}^{\dag,\ddag}$ & 31.5  & $\text{32.5}^{\dag}$  & 31.8 \\
    OPT-1.3B & 27.9  & 27.7  & $\text{28.0}^{\ddag}$ & 8.8   & $\text{13.3}^{\dag}$  & $\text{15.9}^{\dag,\ddag}$ & 32.6  & $\text{33.6}^{\dag}$  & $\text{32.9}^{\dag}$  \\
    OPT-6.7B & 29.6  & $\text{30.7}^{\dag}$  & $\text{30.6}^{\dag}$  & 10.7  & $\text{12.2}^{\dag}$  & $\text{15.0}^{\dag,\ddag}$ & 34.2  & $\text{36.4}^{\dag}$  & $\text{36.9}^{\dag,\ddag}$ \\
    OPT-13B & 27.5  & $\text{29.5}^{\dag}$  & $\text{30.8}^{\dag,\ddag}$ & 9.6   & $\text{11.7}^{\dag}$  & $\text{17.9}^{\dag,\ddag}$ & 31.9  & $\text{35.5}^{\dag}$  & $\text{37.5}^{\dag,\ddag}$ \\
    OPT-66B & 29.5  & $\text{31.0}^{\dag}$  & $\text{33.4}^{\dag,\ddag}$ & 9.1   & $\text{12.1}^{\dag}$  & $\text{16.8}^{\dag,\ddag}$ & 32.1  & $\text{35.3}^{\dag}$  & $\text{36.4}^{\dag,\ddag}$ \\
    \midrule
    \rowcolor{palepink}  \textbf{Avg.} & 28.7  & $\text{29.4}^{\dag}$  & $\text{30.3}^{\dag,\ddag}$ & 10.0  & $\text{12.2}^{\dag}$  & $\text{16.3}^{\dag,\ddag}$ & 32.5  & $\text{34.6}^{\dag}$  & $\text{35.1}^{\dag,\ddag}$ \\
    \midrule
    \textbf{FLAN-T5} &       &       &       &       &       &       &       &       &  \\
    \midrule
    FT5-S & 27.6  & $\text{28.7}^{\dag}$  & 27.0  & 12.6  & 9.4   & $\text{15.0}^{\dag,\ddag}$ & 33.5  & 33.3  & 31.3 \\
    FT5-B & 25.5  & 25.4  & $\text{27.4}^{\dag,\ddag}$ & 10.4  & 10.2  & $\text{15.9}^{\dag,\ddag}$ & 29.8  & 29.6  & $\text{30.0}^{\ddag}$ \\
    FT5-L & 28.5  & 28.5  & $\text{28.8}^{\dag,\ddag}$ & 7.9   & $\text{13.0}^{\dag}$  & $\text{15.6}^{\dag,\ddag}$ & 30.7  & $\text{31.6}^{\dag}$  & $\text{32.1}^{\dag,\ddag}$ \\
    FT5-XL & 28.1  & 27.0  & $\text{28.1}^{\ddag}$ & 9.4   & $\text{10.2}^{\dag}$  & $\text{14.0}^{\dag,\ddag}$ & 30.4  & $\text{33.5}^{\dag}$  & $\text{34.2}^{\dag,\ddag}$ \\
    FT5-XXL & 29.0  & $\text{29.4}^{\dag}$  & $\text{30.5}^{\dag,\ddag}$ & 7.6   & $\text{12.2}^{\dag}$  & $\text{16.2}^{\dag,\ddag}$ & 30.7  & $\text{33.3}^{\dag}$  & $\text{33.8}^{\dag,\ddag}$ \\
    \midrule
     \rowcolor{palepink} \textbf{Avg.} & 27.7  & 27.8  & $\text{28.3}^{\dag,\ddag}$ & 9.6   & $\text{11.0}^{\dag}$  & $\text{15.4}^{\dag,\ddag}$ & 31.0  & $\text{32.3}^{\dag}$  & $\text{32.3}^{\dag}$  \\
    \midrule
   \textbf{Overall Avg} & 27.4  & $\text{27.8}^{\dag}$  & $\text{29.7}^{\dag,\ddag}$ & 10.2  & $\text{11.1}^{\dag}$  & $\text{16.0}^{\dag,\ddag}$ & 31.0  & $\text{32.1}^{\dag}$  & $\text{33.3}^{\dag,\ddag}$ \\
    \bottomrule
    \end{tabular}
    \caption{Spearman correlations on MQM-2020 dataset for machine translation task.
     VAL, IST, and IDM denote the evaluator with vanilla, instruction, and the combination of instruction and demonstration, respectively.
    Values with $\dag$ denote the evaluator with instruction significantly outperforms with vanilla, 
    and values with $\ddag$ denote the evaluator with the combination of instruction and demonstration significantly outperforms with only instruction.
    Values in bold are the best performance in a set of variants (e.g., GPT3 family).
    }
  \label{tab:mqm2020-full}
\end{table*}

\begin{table*}[htb]
  \centering \footnotesize  
    \begin{tabular}{lccccccccc}
    \toprule
    \multirow{2}[2]{*}{\textbf{Model}} & \multicolumn{3}{c}{\textbf{INF}} & \multicolumn{3}{c}{\textbf{NAT}} & \multicolumn{3}{c}{\textbf{FLU}} \\
    \cmidrule(lr){2-4} \cmidrule(lr){5-7} \cmidrule(lr){8-10} 
          & \textbf{VAL} & \textbf{IST} & \textbf{IST+DM} & \textbf{VAL} & \textbf{IST} & \textbf{IST+DM} & \textbf{VAL} & \textbf{IST} & \textbf{IST+DM} \\
    \midrule
    ROUGE-1 & 28.7  & -     & -     & 5.0   & -     & -     & 8.3   & -     & - \\
    ROUGE-2 & 24.0  & -     & -     & 15.2  & -     & -     & 16.0  & -     & - \\
    ROUGE-L & 26.3  & -     & -     & 10.5  & -     & -     & 11.0  & -     & - \\
    BERTScore & 37.2  & -     & -     & 16.0  & -     & -     & 18.7  & -     & - \\
    MoverScore & 30.7  & -     & -     & 20.4  & -     & -     & 14.8  & -     & - \\
    PRISM & 36.8  & -     & -     & 28.7  & -     & -     & 34.4  & -     & - \\
    BARTSCORE & 29.5  & -     & -     & 24.0  & -     & -     & 29.7  & -     & - \\
    +CNN  & 37.7  & -     & -     & 30.1  & -     & -     & 34.4  & -     & - \\
    +CNN+Para & 39.2  & -     & -     & 31.0  & -     & -     & 44.9  & -     & - \\
    \midrule
    \textbf{GPT3} &       &       &       &       &       &       &       &       &  \\
    \midrule
    GPT3-a01 & 33.3  & $\text{37.0}^{\dag}$ & $\text{42.5}^{\dag,\ddag}$ & 20.5  & $\text{28.7}^{\dag}$ & $\text{\textbf{41.7}}^{\dag,\ddag}$ & 28.8  & $\text{\textbf{35.1}}^{\dag}$ & $\text{40.2}^{\dag,\ddag}$ \\
    GPT3-b01 & 39.2  & $\text{\textbf{44.5}}^{\dag}$ & $\text{42.2}^{\dag}$ & 18.2  & $\text{\textbf{29.8}}^{\dag}$ & $\text{39.1}^{\dag,\ddag}$ & 30.0  & $\text{33.8}^{\dag}$ & $\text{40.3}^{\dag,\ddag}$ \\
    GPT3-c01 & 30.6  & $\text{40.9}^{\dag}$ & $\text{\textbf{47.5}}^{\dag,\ddag}$ & 24.8  & $\text{26.5}^{\dag}$ & $\text{39.9}^{\dag,\ddag}$ & 27.4  & $\text{34.2}^{\dag}$ & $\text{44.2}^{\dag,\ddag}$ \\
    GPT3-d01 & \textbf{41.2}  & 39.4  & $\text{43.6}^{\dag,\ddag}$ & \textbf{25.4}  & $\text{26.2}^{\dag}$ & $\text{36.6}^{\dag,\ddag}$ & 29.7  & 27.1  & $\text{\textbf{47.9}}^{\dag,\ddag}$ \\
    GPT3-d03 & 32.9  & 29.8  & $\text{42.0}^{\dag,\ddag}$ & 19.5  & $\text{21.4}^{\dag}$ & $\text{27.5}^{\dag,\ddag}$ & \textbf{36.6}  & 34.2  & $\text{44.4}^{\dag,\ddag}$ \\
    \midrule
   \rowcolor{palepink} \textbf{Avg.} & 35.4  & $\text{38.3}^{\dag}$ & $\text{43.6}^{\dag,\ddag}$ & 21.7  & $\text{26.5}^{\dag}$ & $\text{36.9}^{\dag,\ddag}$ & 30.5  & $\text{32.9}^{\dag}$ & $\text{43.4}^{\dag,\ddag}$ \\
    \midrule
    \textbf{GPT2} &       &       &       &       &       &       &       &       &  \\
    GPT2-M & 39.4  & $\text{42.9}^{\dag}$ & 38.6  & 31.2  & $\text{33.2}^{\dag}$ & $\text{34.3}^{\dag,\ddag}$ & 38.9  & 38.9  & $\text{39.6}^{\dag,\ddag}$ \\
    GPT2-L & 39.7  & $\text{42.2}^{\dag}$ & $\text{41.8}^{\dag}$ & 30.1  & $\text{33.5}^{\dag}$ & $\text{33.1}^{\dag}$ & 34.0  & $\text{40.0}^{\dag}$ & $\text{39.6}^{\dag}$ \\
    GPT2-XL & 41.2  & $\text{42.0}^{\dag}$ & 38.7  & 31.7  & $\text{33.7}^{\dag}$ & $\text{34.8}^{\dag,\ddag}$ & 38.0  & $\text{40.6}^{\dag}$ & $\text{44.2}^{\dag,\ddag}$ \\
    GPT-J-6B & 42.8  & $\text{45.6}^{\dag}$ & 41.6  & 32.5  & 31.5  & $\text{31.9}^{\ddag}$ & 35.9  & $\text{37.7}^{\dag}$ & $\text{42.0}^{\dag,\ddag}$ \\
    \midrule
   \rowcolor{palepink} \textbf{Avg.} & 40.8  & $\text{43.2}^{\dag}$ & 40.2  & 31.4  & $\text{33.0}^{\dag}$ & $\text{33.5}^{\dag,\ddag}$ & 36.7  & $\text{39.3}^{\dag}$ & $\text{41.3}^{\dag,\ddag}$ \\
    \midrule
    \textbf{OPT} &       &       &       &       &       &       &       &       &  \\
    \midrule
    OPT-350M & 37.0  & 36.8  & $\text{37.9}^{\dag,\ddag}$ & 33.9  & 32.5  & 31.1  & 39.9  & 39.5  & $\text{39.9}^{\ddag}$ \\
    OPT-1.3B & 36.7  & $\text{39.3}^{\dag}$ & $\text{38.2}^{\dag}$ & 28.8  & $\text{30.0}^{\dag}$ & $\text{32.9}^{\dag,\ddag}$ & 37.3  & 34.9  & $\text{40.9}^{\dag,\ddag}$ \\
    OPT-6.7B & 40.4  & 39.3  & 38.3  & 31.6  & 27.2  & $\text{35.2}^{\dag,\ddag}$ & 36.0  & 34.4  & $\text{43.6}^{\dag,\ddag}$ \\
    OPT-13B & 37.9  & 37.6  & $\text{38.9}^{\dag,\ddag}$ & 31.4  & 30.3  & $\text{34.6}^{\dag,\ddag}$ & 39.2  & 39.0  & $\text{41.2}^{\dag,\ddag}$ \\
    OPT-66B & 41.4  & $\text{43.2}^{\dag}$& 39.6  & 31.3  & 30.2  & $\text{34.7}^{\dag,\ddag}$ & 36.3  & $\text{37.6}^{\dag}$ & $\text{42.0}^{\dag,\ddag}$ \\
    \midrule
  \rowcolor{palepink}  \textbf{Avg.} & 38.7  & 39.3  & 38.6  & 31.4  & 30.0  & $\text{33.7}^{\dag,\ddag}$ & 37.7  & 37.1  & $\text{41.5}^{\dag,\ddag}$ \\
    \midrule
    \textbf{FLAN-T5} &       &       &       &       &       &       &       &       &  \\
    \midrule
    FT5-S & 39.8  & 37.6  & \text{38.2} & 33.0  & 29.5  & 26.6  & 46.1  & 34.7  & $\text{36.1}^{\ddag}$ \\
    FT5-B & 39.7  & $\text{43.6}^{\dag}$& 37.7  & 26.4  & $\text{30.3}^{\dag}$ & $\text{27.3}^{\dag}$ & 37.8  & $\text{40.6}^{\dag}$ & 37.9 \\
    FT5-L & 42.0  & $\text{42.8}^{\dag}$ & 38.9  & 23.6  & $\text{31.0}^{\dag}$ & $\text{32.6}^{\dag,\ddag}$ & 35.3  & $\text{43.3}^{\dag}$ & $\text{44.5}^{\dag,\ddag}$ \\
    FT5-XL & 41.0  & $\text{42.8}^{\dag}$ & $\text{43.3}^{\dag,\ddag}$ & 24.8  & $\text{28.9}^{\dag}$ & $\text{27.8}^{\dag}$& 37.4  & $\text{44.4}^{\dag}$ & $\text{41.9}^{\dag}$ \\
    FT5-XXL & 44.9  & 40.7  & 37.4  & 24.8  & $\text{28.8}^{\dag}$ & $\text{28.4}^{\dag}$ & 34.2  & $\text{42.5}^{\dag}$ & $\text{41.3}^{\dag}$ \\
    \midrule
  \rowcolor{palepink}  \textbf{Avg.} & 41.5  & 41.5  & 39.1  & 26.5  & $\text{29.7}^{\dag}$ & $\text{28.6}^{\dag}$ & 38.1  & $\text{41.1}^{\dag}$ & $\text{40.3}^{\dag}$ \\
    \midrule
    \textbf{Overall Avg} & 39.1  & $\text{40.6}^{\dag}$ & $\text{40.3}^{\dag}$ & 27.7  & $\text{29.8}^{\dag}$ & $\text{33.2}^{\dag,\ddag}$ & 35.8  & $\text{37.6}^{\dag}$ & $\text{41.6}^{\dag,\ddag}$ \\
    \bottomrule
    \end{tabular}
  \caption{Spearman correlations on BAGEL dataset for data-to-text task.
  VAL, IST, and IDM denote the evaluator with vanilla, instruction, and the combination of instruction and demonstration, respectively.
    Values with $\dag$ denote the evaluator with instruction significantly outperforms with vanilla, 
    and values with $\ddag$ denote the evaluator with the combination of instruction and demonstration significantly outperforms with only instruction.
    Values in bold are the best performance in a set of variants (e.g., GPT3 family).
  }
  \label{tab:d2t-bagel-full}
\end{table*}

\begin{table*}[htb]
  \centering \footnotesize  
    \begin{tabular}{lccccccccc}
    \toprule
    \multirow{2}[2]{*}{\textbf{Model}} & \multicolumn{3}{c}{\textbf{INF}} & \multicolumn{3}{c}{\textbf{NAT}} & \multicolumn{3}{c}{\textbf{FLU}} \\
    \cmidrule(lr){2-4} \cmidrule(lr){5-7} \cmidrule(lr){8-10} 
          & \textbf{VAL} & \textbf{IST} & \textbf{IST+DM} & \textbf{VAL} & \textbf{IST} & \textbf{IST+DM} & \textbf{VAL} & \textbf{IST} & \textbf{IST+DM} \\
    \midrule
    ROUGE-1 & 24.2  & -     & -     & 24.2  & -     & -     & 15.1  & -     & - \\
    ROUGE-2 & 21.9  & -     & -     & 25.9  & -     & -     & 11.4  & -     & - \\
    ROUGE-L & 18.5  & -     & -     & 20.2  & -     & -     & 1.7   & -     & - \\
    BERTScore & 25.8  & -     & -     & 28.0  & -     & -     & 11.8  & -     & - \\
    MoverScore & 17.9  & -     & -     & 24.4  & -     & -     & 5.0   & -     & - \\
    PRISM & 27.4  & -     & -     & 33.1  & -     & -     & 14.2  & -     & - \\
    BARTSCORE & 22.4  & -     & -     & 25.5  & -     & -     & 6.9   & -     & - \\
    +CNN  & 24.2  & -     & -     & 30.6  & -     & -     & 17.2  & -     & - \\
    +CNN+Para & 25.0  & -     & -     & 30.2  & -     & -     & 19.5  & -     & - \\
    \midrule
    \textbf{GPT3} &       &       &       &       &       &       &       &       &  \\
    \midrule
    GPT3-a01 & 25.4  & 19.1  & $\text{25.6}^{\ddag}$ & \textbf{28.7}  & $\text{\textbf{34.0}}^{\dag}$  & $\text{\textbf{37.7}}^{\dag,\ddag}$ & 30.7  & 27.0  & 26.6 \\
    GPT3-b01 & \textbf{37.5}  & 28.4  & 26.5  & 21.5  & $\text{30.6}^{\dag}$  & $\text{26.1}^{\dag}$  & 24.6  & $\text{28.9}^{\dag}$  & 21.1 \\
    GPT3-c01 & 29.8  & 21.3  & $\text{33.7}^{\dag,\ddag}$ & 24.7  & $\text{28.5}^{\dag}$  & $\text{28.6}^{\dag}$  & 31.1  & 27.1  & $\text{27.6}^{\ddag}$ \\
    GPT3-d01 & 32.6  & 27.0  & $\text{33.9}^{\dag,\ddag}$ & 27.3  & $\text{31.7}^{\dag}$  & 21.9  & \textbf{35.8}  & $\text{\textbf{39.7}}^{\dag}$  & 27.1 \\
    GPT3-d03 & 26.6  & $\text{\textbf{29.6}}^{\dag}$  & $\text{\textbf{37.6}}^{\dag,\ddag}$ & 22.6  & $\text{27.0}^{\dag}$  & 18.2  & 33.9  & 31.9  & \textbf{28.2} \\
    \midrule
    \rowcolor{palepink} \textbf{Avg.} & 30.4  & 25.1  & $\text{31.5}^{\dag,\ddag}$ & 25.0  & $\text{30.4}^{\dag}$  & $\text{26.5}^{\dag}$  & 31.2  & 30.9  & 26.1 \\
    \midrule
    \textbf{GPT2} &       &       &       &       &       &       &       &       &  \\
    \midrule
    GPT2-M & 24.7  & 23.1  & 18.2  & 28.7  & $\text{32.7}^{\dag}$  & $\text{35.2}^{\dag,\ddag}$ & 18.7  & $\text{34.8}^{\dag}$  & $\text{33.6}^{\dag}$  \\
    GPT2-L & 19.6  & $\text{28.1}^{\dag}$  & $\text{20.2}^{\dag}$  & 31.2  & $\text{32.4}^{\dag}$  & $\text{37.8}^{\dag,\ddag}$ & 18.6  & $\text{33.1}^{\dag}$  & $\text{35.9}^{\dag,\ddag}$ \\
    GPT2-XL & 22.0  & $\text{23.6}^{\dag}$  & $\text{23.8}^{\dag}$  & 29.7  & 29.1  & $\text{38.0}^{\dag,\ddag}$ & 18.2  & $\text{29.8}^{\dag}$  & $\text{37.1}^{\dag,\ddag}$ \\
    GPT-J-6B & 23.9  & $\text{25.6}^{\dag}$  & 19.6  & 34.3  & 33.3  & $\text{36.8}^{\dag,\ddag}$ & 24.4  & $\text{34.5}^{\dag}$  & $\text{38.4}^{\dag,\ddag}$ \\
    \midrule
    \rowcolor{palepink} \textbf{Avg.} & 22.5  & $\text{25.1}^{\dag}$ & 20.5  & 31.0  & $\text{31.9}^{\dag}$  & $\text{37.0}^{\dag,\ddag}$ & 20.0  & $\text{33.1}^{\dag}$  & $\text{36.2}^{\dag,\ddag}$ \\
    \midrule
    \textbf{OPT} &       &       &       &       &       &       &       &       &  \\
    \midrule
    OPT-350M & 26.1  & $\text{28.7}^{\dag}$  & 25.4  & 27.0  & $\text{29.5}^{\dag}$  & $\text{35.0}^{\dag,\ddag}$ & 21.7  & $\text{26.6}^{\dag}$  & $\text{27.3}^{\dag,\ddag}$ \\
    OPT-1.3B & 26.1  & $\text{28.3}^{\dag}$  & 23.5  & 26.0  & $\text{30.5}^{\dag}$  & $\text{38.7}^{\dag,\ddag}$ & 23.0  & $\text{26.9}^{\dag}$  & $\text{29.8}^{\dag,\ddag}$ \\
    OPT-6.7B & 26.2  & 26.0  & 24.2  & 26.7  & $\text{31.0}^{\dag}$  & $\text{36.5}^{\dag,\ddag}$ & 21.7  & $\text{25.8}^{\dag}$  & $\text{35.9}^{\dag,\ddag}$ \\
    OPT-13B & 27.7  & 26.9  & 26.0  & 24.4  & $\text{30.1}^{\dag}$  & $\text{38.0}^{\dag,\ddag}$ & 20.2  & $\text{29.6}^{\dag}$  & $\text{34.9}^{\dag,\ddag}$ \\
    OPT-66B & 20.1  & $\text{24.7}^{\dag}$  & $\text{22.4}^{\dag}$  & 26.8  & $\text{29.1}^{\dag}$ & $\text{34.6}^{\dag,\ddag}$ & 19.8  & 19.1  & $\text{25.3}^{\dag,\ddag}$ \\
    \midrule
    \rowcolor{palepink} \textbf{Avg.} & 25.2  & $\text{26.9}^{\dag}$  & 24.3  & 26.2  & $\text{30.0}^{\dag}$  & $\text{36.6}^{\dag,\ddag}$ & 21.3  & $\text{25.6}^{\dag}$  & $\text{30.6}^{\dag,\ddag}$ \\
    \midrule
    \textbf{FLAN-T5} &       &       &       &       &       &       &       &       &  \\
    \midrule
    FT5-S & 19.7  & 16.9  & 17.0  & 33.6  & 33.1  & 33.0  & 19.4  & 17.2  & 15.9 \\
    FT5-B & 24.2  & 23.7  & 20.9  & 31.7  & $\text{32.5}^{\dag}$  & $\text{33.4}^{\dag,\ddag}$ & 14.2  & $\text{15.5}^{\dag}$  & $\text{16.8}^{\dag,\ddag}$ \\
    FT5-L & 24.9  & 22.3  & 20.6  & 36.2  & $\text{37.1}^{\dag}$  & $\text{38.6}^{\dag,\ddag}$ & 24.3  & 18.1  & $\text{21.1}^{\ddag}$  \\
    FT5-XL & 26.1  & 23.7  & 19.5  & 38.4  & 35.6  & $\text{37.4}^{\ddag}$  & 28.4  & 21.0  & $\text{22.5}^{\ddag}$  \\
    FT5-XXL & 24.9  & 22.9  & 20.3  & 31.9  & $\text{34.7}^{\dag}$  & $\text{41.7}^{\dag,\ddag}$ & 23.8  & 16.9  & $\text{22.2}^{\ddag}$ \\
    \midrule
    \rowcolor{palepink} \textbf{Avg.} & 24.0  & 21.9  & 19.7  & 34.3  & $\text{34.6}^{\dag}$  & $\text{36.8}^{\dag,\ddag}$ & 22.0  & 17.8  & $\text{19.7}^{\ddag}$ \\
    \midrule
    \textbf{Overall Avg} & 25.5  & 24.7  & 24.0  & 29.1  & 31.7  & $\text{34.2}^{\dag,\ddag}$ & 23.6  & $\text{26.8}^{\dag}$  & $\text{28.2}^{\dag,\ddag}$  \\
    \bottomrule
    \end{tabular}%
   \caption{Spearman correlations on SFRES dataset for data-to-text task.
    VAL, IST, and IDM denote the evaluator with vanilla, instruction, and the combination of instruction and demonstration, respectively.
    Values with $\dag$ denote the evaluator with instruction significantly outperforms with vanilla, 
    and values with $\ddag$ denote the evaluator with the combination of instruction and demonstration significantly outperforms with only instruction.
    Values in bold are the best performance in a set of variants (e.g., GPT3 family).}
  \label{tab:d2t-sfres-full}
\end{table*}

\end{document}